\documentclass{article}
\usepackage{hyperref}
\hypersetup{
  pdftitle={XRM},
  pdfauthor={Pezeshki et al.}
}
\usepackage{microtype}
\usepackage{graphicx}
\usepackage{booktabs}

\usepackage[accepted]{icml2024}
\usepackage{amsmath}
\usepackage{amssymb}
\usepackage{mathtools}
\usepackage{amsthm}
\usepackage[capitalize,noabbrev]{cleveref}
\usepackage{times}
\usepackage{amsfonts}
\usepackage{microtype}
\usepackage{wrapfig}
\usepackage{titlesec}
\usepackage{enumitem}
\usepackage{listings}
\usepackage{multirow}
\usepackage{adjustbox}
\usepackage{subcaption}
\usepackage{colortbl}
\usepackage{stmaryrd}
\usepackage{pifont}
\usepackage{nicematrix}
\usepackage{multicol}
\usepackage{array}
\usepackage{tikz}

\usetikzlibrary{calc, arrows}

\newcommand{\method}{\textsc{XRM}}
\newcommand{\fullmethod}{\textsc{Cross-Risk Minimization}}
\newcommand{\cmark}{\ding{51}}
\newcommand{\xmark}{\ding{55}}
\newcommand{\percent}{\%}
\newcommand{\sqboxs}{1.5ex}

\newcommand{\sqbox}[1]{\textcolor{#1}{\rule{\sqboxs}{\sqboxs}}}
\newcommand{\sqboxblack}[1]{\setlength{\fboxsep}{0pt}\fbox{\sqbox{#1}}}

\definecolor{papercolor}{HTML}{0668E1}
\definecolor{windows_98}{HTML}{008080}

\hypersetup{
    colorlinks,
    linkcolor={papercolor!75!black},
    citecolor={papercolor!75!black},
    urlcolor={papercolor!75!black}
}

\theoremstyle{plain}

\theoremstyle{definition}

\theoremstyle{remark}

\usepackage[textsize=tiny]{todonotes}

\icmltitlerunning{Discovering Environments with \method{}}

\begin{document}

\twocolumn[
\icmltitle{Discovering Environments with \method{}}

\begin{icmlauthorlist}
\icmlauthor{Mohammad Pezeshki}{fair}
\icmlauthor{Diane Bouchacourt}{fair}
\icmlauthor{Mark Ibrahim}{fair}
\icmlauthor{Nicolas Ballas}{fair}\\
\icmlauthor{Pascal Vincent}{fair,mila,cifar}
\icmlauthor{David Lopez-Paz}{fair}
\end{icmlauthorlist}

\icmlaffiliation{fair}{FAIR at Meta}
\icmlaffiliation{mila}{Mila at Université de Montréal}
\icmlaffiliation{cifar}{CIFAR}

\icmlcorrespondingauthor{Mohammad Pezeshki}{mpezeshki@meta.com}
\vspace{-0.4cm}

\icmlkeywords{Machine Learning, ICML}

\vskip 0.3in
]

\printAffiliationsAndNotice{}

\begin{abstract}
  Environment annotations are essential for the success of many out-of-distribution (OOD) generalization methods.
  Unfortunately, these are costly to obtain and often limited by human annotators' biases.
  To achieve robust generalization, it is essential to develop algorithms for automatic environment discovery within datasets.
  Current proposals, which divide examples based on their training error, suffer from one fundamental problem.
  These methods introduce hyper-parameters and early-stopping criteria, which require a validation set with human-annotated environments, the very information subject to discovery.
  In this paper, we propose \fullmethod{} (\method{}) to address this issue.
  \method{} trains twin networks, each learning from one random half of the training data, while imitating confident held-out mistakes made by its sibling.
  \method{} provides a recipe for hyper-parameter tuning, does not require early-stopping, and can discover environments for all training and validation data.
  Algorithms built on top of \method{} environments achieve oracle worst-group-accuracy, addressing a long-standing challenge in OOD generalization.
  Code available at \url{https://github.com/facebookresearch/XRM}.
\end{abstract}

\vspace{-0.9cm}
\section{Introduction}
\label{sec:intro}
AI systems pervade our lives, spanning applications such as finance~\citep{ai_finance}, healthcare~\citep{ai_healthcare}, self-driving vehicles~\citep{ai_driving}, and justice~\citep{ai_justice}.
Despite outperforming humans in many tasks, these systems fall apart under testing conditions different from their \emph{training environments}~\citep{geirhos2020shortcut}.
For instance, during the COVID-19 pandemic, thoracic x-ray classifiers incorrectly latched onto spurious correlations such as patient's age or position~\citep{covid2}, leading to ``an alarming situation in which the systems appear accurate, but fail when tested in new hospitals''~\citep{covid}.

\begin{figure*}
    \centering
    \includegraphics[width=0.8\textwidth]{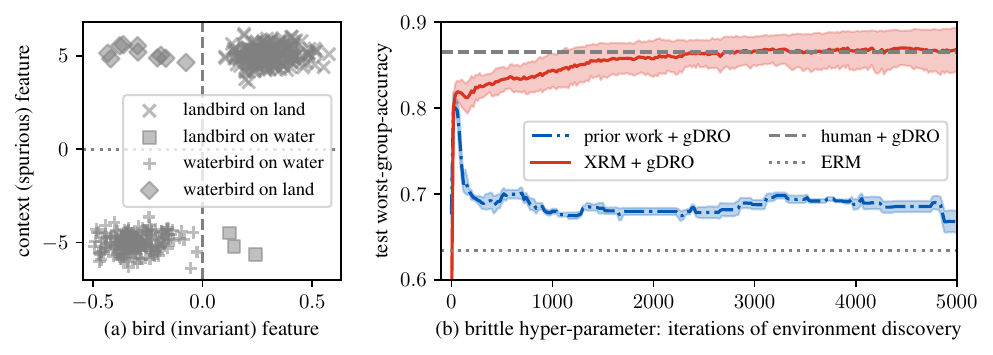}
    \caption{
    (a) Waterbirds problem with four groups: a \emph{majority group} of waterbirds in water, landbirds in land, waterbirds in land, and a \emph{minority group} of landbirds in water.
    Models often rely on spurious features to classify the majority of examples and then memorize the minority examples.
    (b) Worst-group-accuracy (minority) for different methods.
    (Dotted line) ERM achieves $61\percent{}$.
    (Dashed line) GroupDRO with human group annotations (oracle) achieves $87\percent{}$.
    (Dashdot blue line) Prior work to discover groups requires early-stopping with surgical precision.
    (Solid red line) \method{} enables an oracle performance of $87\percent{}$ without requiring early stopping.
    }
    \vspace{-0.4cm}
    \label{fig:one}
\end{figure*}

Generally speaking, AI systems underperform on under-represented groups in training data~\citep{barocas-hardt-narayanan}.
The Waterbirds problem~\citep{dro}, depicted in \cref{fig:one}, illustrates this with two classes (landbirds and waterbirds) in two landscape environments (land and water), forming four groups: a \emph{majority group} of waterbirds in water (73\percent{} of examples), landbirds in land (22\percent{}), waterbirds in land (4\percent{}), and a \emph{minority group} of landbirds in water (1\percent{}).
On this problem, learning machines often favor the \emph{landscape} spurious feature to classify the majority of examples and memorizes the remaining minorities to achieve zero training error.
An empirical risk minimization (ERM) baseline, ignoring environment data~\citep{vapnik}, achieves a mere 61\percent{} worst-group-accuracy, specifically on the minority group, as illustrated in \cref{fig:one}'s right panel.
    
To improve upon ERM, researchers have developed a myriad of OOD generalization algorithms~\citep{dg_survey, dg_survey_2}.
These methods use environment annotations to uncover invariant (environment-generic) patterns and discard spurious (environment-specific) ones~\citep{irm}.
As~\cref{fig:one} shows, \emph{group distributionally robust optimization}~\citep[GroupDRO]{dro} achieves a worst-group-accuracy of $87\percent{}$.
This outperforms ERM by over twenty five points, a sizeable gap!

While promising, OOD algorithms targeting sub-population shift require environment annotations, which are costly to obtain and limited by human annotators' biases and precision.
Moreover, no single set of environment annotations fits all OOD algorithms. The patterns misleading a learning system might be alien or invisible to humans~\citep{adversarial}.
Because of these reasons, OOD generalization is currently confined to small data collections, and their promise in the large-scale setting remains unfulfilled.

In light of the above, researchers developed algorithms for automatic environment discovery~\citep{ls, jtt, cnc, arl, toogood, eiil, lff}.
These methods typically build a robust system in two phases.
In phase-1, these methods train a classifier to categorize training examples in two environments, based on their training error.
In phase-2, an OOD algorithm is trained on top of the discovered environments.
However, this pipeline suffers from one fundamental issue: the need for precisely controlling the classifier's capacity, ensuring that the discovered environments differ only in spurious correlations.
As \cref{fig:one} shows, peak performance (79\percent{}) is reached with exact early-stopping in phase-1; without it, accuracy drops to 68\percent{}.
Lacking a precise indicator of early-stopping, environment discovery methods depend on validation sets with human annotations.
Alas, at least in our view, this defeats the \emph{raison d'\^{e}tre} of environment discovery.

\vspace{-0.2cm}
\paragraph{Contribution}
We propose \fullmethod{}~(\method{}), a simple method for environment discovery that requires no human environment annotations whatsoever. 
\method{} trains two twin networks, each holding-in one random half of the training data.
During training, \method{} instructs each twin to imitate confident held-out mistakes made by their sibling.
This results in an ``echo-chamber'' where twins increasingly rely on bias, converging on a pair of environments that differ in spurious correlation, and share the invariances that fuel downstream out-of-distribution generalization.
After twin training, a simple cross-mistake formula allows \method{} to annotate all of the training and validation examples with environments.
As our experiments show, \method{} endows OOD generalization algorithms with oracle-like performance across benchmarks.
Returning one final time to~\cref{fig:one}, we observe that \method{}+GroupDRO converges to $87\percent{}$ worst-group-accuracy on Waterbirds, matching the oracle!

The sequel is as follows.
Section~\ref{sec:invariance_learning} details the problem formulation.
Section~\ref{sec:related_main} surveys prior research on environment discovery.
Section~\ref{sec:xrm} describes \method{}.
Section~\ref{sec:experiments} demonstrates the effectiveness of \method{}, while Section~\ref{sec:fail} discusses its potential shortcomings. Section~\ref{sec:discussion} concludes with thoughts for future work.

\section{Learning Invariances Across Environments}
\label{sec:invariance_learning}

The goal of OOD generalization is to build learning systems that perform well beyond the training data distribution.
To this end, we collect examples from multiple environments and
OOD algorithms search for invariant patterns across these environments, while disregarding environment-specific spurious correlations~\citep{irm}.
Formally, we seek a predictor $f$ that classifies inputs $x$ into labels $y$, across all relevant environments $e \in \mathcal{E}$:
\begin{equation}\label{eq:ood}
    f \in \mathop{\mathrm{argmin}}_{\tilde f} \sup_{e \in \mathcal{E}} R^e(\tilde f),
\end{equation}
where the risk $R^e(f) = \mathbb{E}_{(x, y) \sim P^e}[\ell(f(x), y)]$ measures the average loss $\ell$ incurred by the predictor $f$ across examples from environment $e$, all of them drawn iid from $P^e$.

In its full generality, OOD generalization in~\eqref{eq:ood} is an admittedly daunting task.
To alleviate the burden, prior literature often considers the simplified and more practical version of \emph{sub-population shift}~\citep{dro}.
Given a dataset $\mathcal{D} = \{(x_i, y_i, e_i)\}_{i=1}^n$, the supremum in~\eqref{eq:ood} is replaced by a maximum over the training environments and the risk for each environment is approximated by the empirical risk~\citep{vapnik}.
The effectiveness of an OOD algorithm is then assessed by its worst-group-accuracy on a validation set.

In practice, several OOD algorithms have been successful in learning invariances across environments ~\citep{domainbed, dg_survey, dg_survey_2, yang2023change}.
Despite their promise, their large-scale application is hindered by the need for human-annotated environments, which are resource-intensive and might be even sub-optimal.
Different machine learning models fall prey to different kinds of spurious correlations.
In addition, there exists complex interactions between environment definition, function class, distributional shift, and cultural viewpoints~\citep{lopez2022measuring}.
Therefore, environment annotations are helpful only when revealing spurious and invariant patterns under the lens of the learning system under consideration.
Could it be possible to design algorithms for the automatic discovery of environments tailored to the learning machine and data at hand?

\subsection{Discovering Environments}
\label{sec:discovery}
\begin{flushright}
\emph{Nature does not shuffle data}---\citet{bottou_shuffle}
\end{flushright}

\vspace*{-0.4cm}
Let us reconsider the problem of OOD generalization without access to environment annotations.
This time, it suffices to talk about one training distribution $P^\text{tr}$ and one testing distribution $P^\text{te}$.
Our training data is a collection of input-label pairs $(x_i, y_i)$, each drawn iid from the training distribution.
While $P^\text{tr}$ may be the mixture of multiple environments describing interesting invariant and spurious correlations, this rich heterogeneity is shuffled together and unbeknown to us.
But, if we could ``unshuffle'' the training distribution and recover the environments therein, we could invoke the OOD generalization machinery from the previous section and hope for a robust predictor.
This is the purpose of automatic environment discovery.

\section{Related Work on Environment Discovery}
\label{sec:related_main}
To discover environments, prior work often train a classifier and then assign each training example to two environments based on their loss or classification accuracy.
Crucially, one must control the capacity of the classifier with surgical precision, such that it relies only on the spurious correlations.
It is only in such cases that the subsequent OOD generalization algorithms can successfully disregard these spurious features.

As a result, proposals for environment discovery differ mainly in how to control the capacity of the classifier.
For example, the too-good-to-be-true prior~\citep{toogood} employs a classifier with a small parameter count while correct-n-contrast~\citep[CnC]{cnc} applies strong weight decay regularization.
Just train twice~\citep[JTT]{jtt} and environment inference for invariant learning~\citep[EIIL]{eiil} train a classifier for a limited number of epochs.
Learning from failure~\citep[LfF]{lff} biases the classifier towards the use of ``simple'' features by applying a generalized version of the cross entropy loss.
Other proposals, such as learning to split~\citep[LS]{ls} and adversarial re-weighted learning~\citep[ARL]{arl} complement capacity control with adversarial games.

However, all these methods assume having access to a human-annotated validation set to conduct such precise capacity controls. This defeats the purpose of environment discovery.
In fact, if we have access to a small dataset with human-annotated environments, these examples suffice to fine-tune the last layer of a deep network towards state-of-the-art worst-group-accuracy~\citep{izmailov2022feature}.

For a more detailed discussion and related work, please refer to Appendix \ref{app:related_work}.

\section{Cross-Risk Minimization (\method{})}
\label{sec:xrm}

We propose \fullmethod{}~(\method{}), an algorithm to discover environments without the need of human supervision.
\method{} comes with batteries included, namely a recipe for hyper-parameter tuning and a formula to annotate all training and validation data.
As we will show in~\cref{sec:experiments}, environments discovered by \method{} endow OOD generalization algorithms with oracle performance.

The blueprint for \method{} is as follows.
\method{} trains two twin classifiers, each holding-in one random half of the training data~(\cref{sec:twin}).
During training, \method{} biases each twin to absorb spurious correlation by imitating confident held-out mistakes from their sibling~(\cref{sec:flips}).
\method{} chooses hyper-parameters for the twins based on the number of imitated mistakes~(\cref{sec:ourparams}).
Finally, and given the selected twins, \method{} employs a simple ``cross-mistake'' formula to discover environment annotations for all of the training and validation examples~(\cref{sec:crossmistake}).
\Cref{alg:xrm} serves as a companion to the descriptions below; \cref{app:code} contains a PyTorch implementation.
The runtime of \method{} is akin to one ERM baseline on the training data.

\begin{algorithm}
  \caption{\fullmethod{} (\method{})}
  \label{alg:xrm}
  \textbf{Input:} training examples $\{(x_i, y_i)\}_{i=1}^n$ and validation examples $\{(\tilde{x}_i, \tilde{y}_i)\}_{i=1}^m$\\
  \textbf{Output:} discovered environments $\{e_i\}_{i=1}^n$ and $\{\tilde{e}_i\}_{i=1}^m$
  \begin{itemize}[leftmargin=0.5cm,itemsep=0pt]
  \item Fix held-in training example assignments $m^a_i \sim \text{Bernoulli}(\frac{1}{2})$ and $m^b_i = 1 - m^a_i$
  \item Initialize two classifiers $f^a$ and $f^b$
  \item Until convergence: 
  \begin{itemize}
    \item Compute held-in softmax predictions:\\$p^\text{in}_i = m^a_i f^a(x_i) + m_i^b f^b(x_i)$
    \item Compute held-out softmax predictions:\\$p^\text{out}_i = m^b_i f^a(x_i) + m^a_i f^b(x_i)$
    \item Update $f^a$ and $f^b$ to minimize the held-in class-balanced cross-entropy loss $\ell(p^\text{in}, y)$
    \item Flip $y_i$ into $y^\text{out}_i = \mathrm{argmax}_j p^\text{out}_{i,j}$, with probability:\\$(p^\text{out}_{i,y^\text{out}_i} - 1/ n_\text{classes}) \cdot n_\text{classes} / (n_\text{classes} - 1)$
  \end{itemize}
  \item Define cross-mistake function $e(x, y) = \llbracket (y \notin \mathrm{argmax}_j f^a(x)_j) \, \lor \, (y \notin \mathrm{argmax}_j f^b(x)_j) \rrbracket$
  \item Discover training $e_i = e(x_i, y_i)$ and validation $\tilde{e}_i = e(\tilde{x}_i, \tilde{y}_i)$ environments
  \end{itemize}
\end{algorithm}

\subsection{Twin Setup, Holding-out of Data}
\label{sec:twin}

We start by initializing two twin classifiers $f^a$ and $f^b$.
Without loss of generality, let these classifiers return softmax probability vectors over the $n_\text{classes}$ classes in the training data.
We split our training dataset $\{(x_i, y_i)\}_{i=1}^n$ in two random halves.
Formally, we construct a pair of training assignment vectors with entries $m^a_i \sim \text{Bernoulli}(\frac{1}{2})$ and $m^b_i = 1 - m^a_i$, for all $i = 1, \ldots, n$.
For classifier $f^a$, examples with $m^a_i = 1$ are ``held-in'' and examples with $m^a_i = 0$ are ``held-out''; similarly for $f^b$.
Therefore, we will train classifier $f^a$ on training examples where $m^a_i = 1$, and similarly for classifier $f^b$.
See~\cref{app:code} for implementation details.

By virtue of this arrangement, we may now estimate the generalization difficulty of any example by looking at the prediction of the twin that held-out such point.
This contrasts prior methods, which consume the entire training data, and may therefore conflate generalization and memorization.
Here, however, if a point is misclassified when held-out, we see this as evidence of such example belonging to the minority group.
\citet{feldman} proposes a similar ``error when holding-out'' construction as a measure of memorization. In the context of label-noise robustness, CrossSplit~\citep{kim2023crosssplit} also employs a similar approach, in which, confident held-out mistakes are indicators of a model’s memorization of a noisy label.

\begin{table*}[th]
\caption{Worst-group-accuracies, averaged from ten runs across datasets and algorithms, show XRM achieving oracle-level performance. When group labels are not available, class labels substitute them. Additionally, while ERM does not use group labels, it can still benefit from validation group labels for hyperparameter tuning, resulting in improved performance.}
\label{tab:xrm_effectiveness}
\begin{center}
\resizebox{\textwidth}{!}{%
\begin{NiceTabular}{lcccccccccccc}
\CodeBefore
\rectanglecolor{papercolor!10}{2-4}{11-4}
\rectanglecolor{papercolor!10}{2-7}{11-7}
\rectanglecolor{papercolor!10}{2-10}{11-10}
\rectanglecolor{papercolor!10}{2-13}{11-13}
\Body
\toprule
&
\multicolumn{3}{c}{\textbf{ERM}} &
\multicolumn{3}{c}{\textbf{GroupDRO}} &
\multicolumn{3}{c}{\textbf{RWG}} &
\multicolumn{3}{c}{\textbf{SUBG}} \\
\midrule
	&	None	&	Human	&	\textbf{XRM}	&	None	&	Human	&	\textbf{XRM}	&	None	&	Human	&	\textbf{XRM}	&	None	&	Human	&	\textbf{XRM}\\
\midrule																									
\textbf{Waterbirds}     & 66.4 & 66.4 & 66.4 & 67.3 & 86.5 & 88.1 & 66.8 & 87.3 & 82.0 & 62.1 & 87.1 & 77.7 \\
\textbf{CelebA}         & 54.3 & 55.1 & 58.6 & 68.4 & 88.3 & 89.1 & 64.7 & 82.6 & 81.1 & 65.9 & 83.9 & 81.1 \\
\textbf{MultiNLI}       & 67.9 & 72.0 & 69.1 & 68.6 & 73.4 & 72.1 & 66.9 & 71.3 & 70.2 & 69.7 & 52.4 & 72.0 \\
\textbf{CivilComments}  & 67.2 & 74.0 & 64.7 & 66.6 & 73.8 & 72.2 & 67.2 & 73.4 & 72.2 & 65.4 & 71.1 & 44.6 \\
\textbf{ColorMNIST}     & 10.0 & 10.1 & 11.3 & 9.9  & 10.1 & 69.7 & 10.0 & 10.3 & 69.4 & 10.0 & 9.9  & 65.2 \\
\textbf{MetaShift}      & 64.2 & 69.8 & 71.7 & 72.5 & 80.3 & 77.5 & 72.3 & 76.3 & 74.2 & 69.7 & 77.0 & 77.6 \\
\textbf{ImagenetBG}     & 78.4 & 78.4 & 79.0 & 78.3 & 78.2 & 78.2 & 78.3 & 79.0 & 78.8 & 79.5 & 78.8 & 77.2 \\
\midrule
\textbf{Average}        & 58.3 & 60.8 & 60.1 & 61.6 & 70.1 & 78.1 & 60.9 & 68.6 & 75.4 & 60.3 & 65.8 & 70.8 \\
\bottomrule
\end{NiceTabular}%
}
\end{center}
\end{table*}

\subsection{Twin Training, Flipping Labels}
\label{sec:flips}

As \cref{fig:one} shows, the test worst-group-accuracy of an ERM baseline on Waterbirds is $62\percent{}$.
This suggests that, if using ERM to train our twins, each would be able to correctly classify roughly one half of the minority examples.
If using these machines to discover environments based on prediction errors, we would dilute the spurious correlation evenly across the two discovered environments.
Consequently, it would be difficult for an OOD generalization algorithm to tell apart between invariant and spurious patterns.
Albeit counter-intuitive, we would like to hinder the learning process of our twins, such that they increasingly rely on spurious correlation.
In the best possible case, the twins would correctly classify all majority examples and misclassify all minority examples, resulting in \emph{zero} worst-group accuracy.

To this end, we propose to steer away our twins from becoming empirical risk minimizers as follows.
Let $p^\text{out}_i = m^b_i f^a(x_i) + m^a_i f^b(x_i)$ be the held-out softmax prediction for example $(x_i, y_i)$.
Also, let $y^\text{out}_i = \arg\max_j p^\text{out}_{i, j}$ be the held-out predicted class label.
Then, at each iteration during the training of the twins, flip $y_i$ into $y_i^{\text{out}}$, with probability,
\begin{equation}\label{eq:flip}
 (p^{\text{out}}_{y^\text{out}_i} - 1/n_\text{classes}) \cdot n_\text{classes} / (n_\text{classes} -1),
\end{equation}
and let each network to minimize their held-in cross-entropy loss---according to the moving targets.

The overarching intuition is that the label flipping~\cref{eq:flip} implements an ``echo chamber'' reinforcing the twins to rely on spurious correlation.
Label flipping happens more often for confident held-out mistakes and early in training.
These are two footprints of spurious correlations, since these are often easier and faster to capture.
(In the context of neural networks, this is often referred to as a ``simplicity bias''~\citep{simplicity,gs}.)
Overall, the purpose of~\cref{eq:flip} is to transform the labels of the training data such that they no longer represent the original classes, but spurious bias.
Finally, the adjustment of~\cref{eq:flip} in terms of $n_\text{classes}$ ensures low flip probabilities at initialization, where mistakes are random, and not due to spurious correlation.
We note that the ``echo chamber'' effect aligns the twin networks and that is crucially different from methods that use multiple networks to either disagree with or diversify spurious features~\citep{lff, cha2021swad, rame2022diverse, wortsman2022model, lee2022diversify, pagliardini2022agree, lin2023spurious, eastwood2023spuriosity}.

\subsection{Twin Model Selection, Counting Label Flips}
\label{sec:ourparams}

Before discovering environments, we must commit to a pair of twin classifiers.
Each of the twin networks own hyper-parameters, \method{} would be incomplete without a model selection criterion~\citep{domainbed}.
We propose to select the twin hyper-parameters showing a maximum number of label flips at the last iteration, and across the training data.
To reiterate, by ``counting flips'' we simply compare the vector of current labels with the vector of original labels---therefore, we do not accumulate counts of double or multiple flips per label. 
To understand why, recall that each label flip signifies one example that is confidently misclassified when held-out.
Therefore, each label flip is evidence about reliance on spurious correlation, which consequently brings us closer to a clear-cut identification of the minority group.

\subsection{The Cross-Mistake Formula}
\label{sec:crossmistake}

Having committed to a pair of twins, we are ready to discover environments for all of our training and validation examples.
In particular, we use a simple ``cross-mistake'' formula to annotate any example $(x, y)$ with the binary annotation, $e(x, y) =$
\begin{equation}\label{eq:envs}
  \llbracket (y \notin \mathrm{argmax}_j f^a(x)_j) \, \lor \, (y \notin \mathrm{argmax}_j f^b(x)_j) \rrbracket,
\end{equation}
where ``$\lor$'' denotes logical-OR, and ``$\llbracket\,\rrbracket$'' is the Iverson bracket.
If operating within the group-shift paradigm, we define one group per combination of label and discovered environment.
Notably, the ability to annotate both training and validation examples is a feature inherited from holding-out data during twin training.
More particularly, every example---within training and validation sets---is held-out for at least one of the two twins, as subsumed in \cref{eq:envs} by the logical-OR operation.

We are now ready to train the OOD generalization algorithm of our choice on top of the training data with environments discovered with \method{}.

\section{Experiments}
\label{sec:experiments}
This section presents a series of experiments to showcase the effectiveness of \method{} on two well-known benchmarks.
Additional experiments are also conducted to identify scenarios where XRM excels, as well as scenarios where it fails to discover relevant environments.

\subsection{Sub-population Shift Benchmarks}
For sub-population shift tasks, we experiment with seven datasets and four algorithms detailed in Appendix \ref{app:exp}.
We compare results with 3 sources of environment annotations:
\vspace{-0.7cm}
\begin{itemize}[leftmargin=*, itemsep=-3pt]
    \item None: no env. annotations---class labels are used instead,
    \item Human: original human-annotated environments,
    \item XRM: inferred environments discovered by our method.
\end{itemize}

\vspace{-0.1cm}
\paragraph{Metrics}
Regardless of how training and validation environments are discovered, we always report test worst-group-accuracy over the human environment annotations provided by each dataset.
The tables hereby presented show averages over ten random seeds.
For results with error bars, see~\cref{tab:xrm_full}.

\paragraph{\method{} vs. human annotations}

\Cref{tab:xrm_effectiveness} shows that \method{} enables oracle-like worst-group-accuracy across datasets.
The performance gains are remarkable in the challenging ColorMNIST dataset, where \method{} perfectly identifies digits appearing in minority colors, discovering a pair of environments conducive of stronger generalization than the ones originally proposed by humans.
For the commonly-reported quartet of Waterbirds, CelebA, MultiNLI, and CivilComments, the average worst-group-accuracy is $67.3\percent{}$ when no group annotations are used.
When using XRM, the average worst-group-accuracy significantly improves to $80.4\percent{}$, closely matching $80.6\percent{}$ achieved with human annotations.

\paragraph{\method{} vs. other methods}

\Cref{tab:xrm_vs_others} compares the worst-group-accuracy achieved by GroupDRO using XRM-inferred environments against other environment discovery methods.
These include learning from failure~\citep[LfF]{lff}, environment inference for invariant learning~\citep[EIIL]{eiil}, just train twice~\citep[JTT]{jtt}, correct-n-contrast~\citep[CnC]{cnc}, automatic feature re-weighting~\citep[AFR]{qiu2023simple}, and LS~\citep{ls}.
As seen in the previous subsection, \method{} achieves $80.4\percent{}$, nearly matching oracle performance.
The second best method with no access to environment information, JTT, drops to $58.9\percent{}$.
The best method accessing a validation set with human environment annotations, AFR, lags far from \method{}, with $78.6\percent{}$.
The computational burden to complete the results from LS was prohibitive, with more details provided in Appendix \ref{app:ls}.
For example, one run of LS for Waterbirds, the smallest dataset, took 20 hours.
An \method{} run for this same dataset, on the same 32GB Volta GPU, takes 10 minutes.

\begin{table*}
\caption{Average/worst accuracies comparing methods for environment discovery. We specify access to annotations in training data ($e^\text{tr}$) and validation data ($e^\text{va}$). Symbol $\dagger$ denotes original numbers.}
\vspace{-0.4cm}
\label{tab:xrm_vs_others}
\begin{center}
\resizebox{0.85\textwidth}{!}{
\begin{NiceTabular}{lllcccccccc|cc}
\CodeBefore
\rectanglecolor{papercolor!10}{2-5}{17-5}
\rectanglecolor{papercolor!10}{2-7}{17-7}
\rectanglecolor{papercolor!10}{2-9}{17-9}
\rectanglecolor{papercolor!10}{2-11}{17-11}
\rectanglecolor{papercolor!10}{2-13}{17-13}
\Body
\toprule
\multicolumn{2}{c}{} &
\multicolumn{1}{c}{} &
\multicolumn{2}{c}{\textbf{Waterbirds}} &
\multicolumn{2}{c}{\textbf{CelebA}} &
\multicolumn{2}{c}{\textbf{MNLI}} &
\multicolumn{2}{c}{\textbf{CivilComms}} &
\multicolumn{2}{c}{\textbf{Average}}\\
\midrule
$e^\text{tr}$  &  $e^\text{va}$  &    &  Avg  &  Worst  &  Avg  &  Worst  &  Avg  &  Worst  &  Avg  &  Worst  &  Avg  &  Worst  \\
\midrule                                                  
\multirow{2}{*}{\cmark} &   \multirow{2}{*}{\cmark}
            &   ERM             &  83.8 & 66.4  &  95.5 & 55.1  &  81.6 & 72.0  &  84.3 & 74.0  &  86.3 & 66.9  \\
    &       &   GroupDRO        &  90.2 & 86.5  &  93.1 & 88.3  &  80.6 & 73.4  &  84.2 & 73.8  &  87.0 & 80.5  \\
\midrule\multirow{5}{*}{\xmark} &   \multirow{5}{*}{\cmark}
            &   ERM$^\dagger$   &  97.3 & 72.6  &  95.6 & 47.2  &  82.4 & 67.9  &  83.1 & 69.5  &  89.6 & 64.3 \\
    &       &   LfF$^\dagger$   &  91.2 & 78.0  &  85.1 & 77.2  &  80.8 & 70.2  &  68.2 & 50.3  &  81.3 & 68.9 \\
    &       &   EIIL$^\dagger$  &  96.9 & 78.7  &  89.5 & 77.8  &  79.4 & 70.0  &  90.5 & 67.0  &  89.1 & 73.4 \\
    &       &   JTT$^\dagger$   &  93.3 & 86.7  &  88.0 & 81.1  &  78.6 & 72.6  &  83.3 & 64.3  &  85.8 & 76.2 \\
    &       &   CnC$^\dagger$   &  90.9 & 88.5  &  89.9 & 88.8  &  ---  & ---   &  ---  & ---   &  ---  & ---  \\
    &       &   AFR$^\dagger$   &  94.4 & 90.4  &  91.3 & 82.0  &  81.4 & 73.4  &  89.8 & 68.7  &  89.2 & 78.6 \\
\midrule
\multirow{7}{*}{\xmark} &   \multirow{7}{*}{\xmark}
            &   ERM             &  83.5 & 66.4  &  95.4 & 54.3  &  82.1 & 67.9  &  81.3 & 67.2  &  85.6 & 63.9  \\
    &       &   LfF$^\dagger$   &  86.6 & 75.0  &  81.1 & 53.0  &  71.4 & 57.3  &  69.1 & 42.2  &  77.1 & 56.9 \\
    &       &   EIIL$^\dagger$  &  90.8 & 64.5  &  95.7 & 41.7  &  80.3 & 64.7  &  ---  & ---   &  ---  & ---  \\
    &       &   JTT$^\dagger$   &  88.9 & 71.2  &  95.9 & 48.3  &  81.4 & 65.1  &  79.0 & 51.0  &  86.3 & 58.9 \\
    &       &   LS$^\dagger$    &  91.2 & 86.1  &  87.2 & 83.3  &  78.7 & 72.1  &  ---  & ---   &  ---  & ---  \\
    &       &   BAM$^\dagger$   &  91.4 & 89.1  &  88.4 & 80.1  &  80.3 & 70.8  &  88.3 & 79.3  &  87.1 & 79.8 \\
    &       &  XRM              &  89.3 & 88.1  &  91.4 & 89.1  &  75.8 & 72.1  &  84.0 & 72.2  &  85.1 & 80.4  \\

\bottomrule
\end{NiceTabular}%
}
\end{center}
\end{table*}

\subsection{The DomaninBed Benchmark}
\Cref{tab:domainbed_summary} presents additional domain generalization results on the \textsc{DomainBed} benchmark~\citep{domainbed}. 
Experiments compare three settings: ERM without any environment annotations, the CORAL domain generalization algorithm~\citep{sun2016deep} with human-annotated environments, and CORAL with environments discovered by \method{}. 
As a note, CORAL is the best performing single-model (non-ensembling) method in the DomainBed suite.
Once again, results suggest that the performance when using \method{}-inferred annotations is comparable to that of human-annotated environments.
Further details on these experiments are provided in~\cref{app:domainbed} with full table of results in~\cref{tab:domainbed_full}.
\begin{table*}
\caption{The average and worst test environment accuracies for five datasets in the \textsc{DomainBed} benchmark~\citep{domainbed}. Three methods are compared: 1) ERM with no environment annotations, 2) CORAL with human-annotated environments, and 3) CORAL with \method{}-inferred environments. The model selection is done according to the average accuracy over validation environments.}
\vspace{-0.2cm}
\label{tab:domainbed_summary}
\begin{center}
\resizebox{0.9\textwidth}{!}{%
\begin{NiceTabular}{lcccccccccc}
\CodeBefore
\rectanglecolor{papercolor!10}{2-3}{5-3}
\rectanglecolor{papercolor!10}{2-5}{5-5}
\rectanglecolor{papercolor!10}{2-7}{5-7}
\rectanglecolor{papercolor!10}{2-9}{5-9}
\rectanglecolor{papercolor!10}{2-11}{5-11}
\Body
\toprule
&
\multicolumn{2}{c}{\textbf{VLCS}} &
\multicolumn{2}{c}{\textbf{PACS}} &
\multicolumn{2}{c}{\textbf{OfficeHome}} &
\multicolumn{2}{c}{\textbf{TerraInc}} &
\multicolumn{2}{c}{\textbf{DomainNet}} \\
\midrule
\textbf{Method} (annotations) & Avg & Worst & Avg & Worst & Avg & Worst & Avg & Worst & Avg & Worst \\
\midrule
\textbf{ERM} (None)    & 77.97 & 64.85 & 83.35 & 72.55 & 65.47 & 52.25 & 47.02 & 34.60 & 31.69 & 9.30 \\
\textbf{CORAL} (Human) & 77.87 & 65.00 & 84.99 & 77.70 & 67.74 & 53.55 & 48.51 & 37.15 & 41.97 & 13.25 \\
\textbf{CORAL} (XRM)   & 77.66 & 66.15 & 83.81 & 77.30 & 67.01 & 53.90 & 49.60 & 38.00 & 35.87 & 11.60 \\
\bottomrule
\end{NiceTabular}%
}
\end{center}
\end{table*}

\subsection{Further Analytical Experiments}

\begin{figure}[t!]
\includegraphics[width=0.5\textwidth]{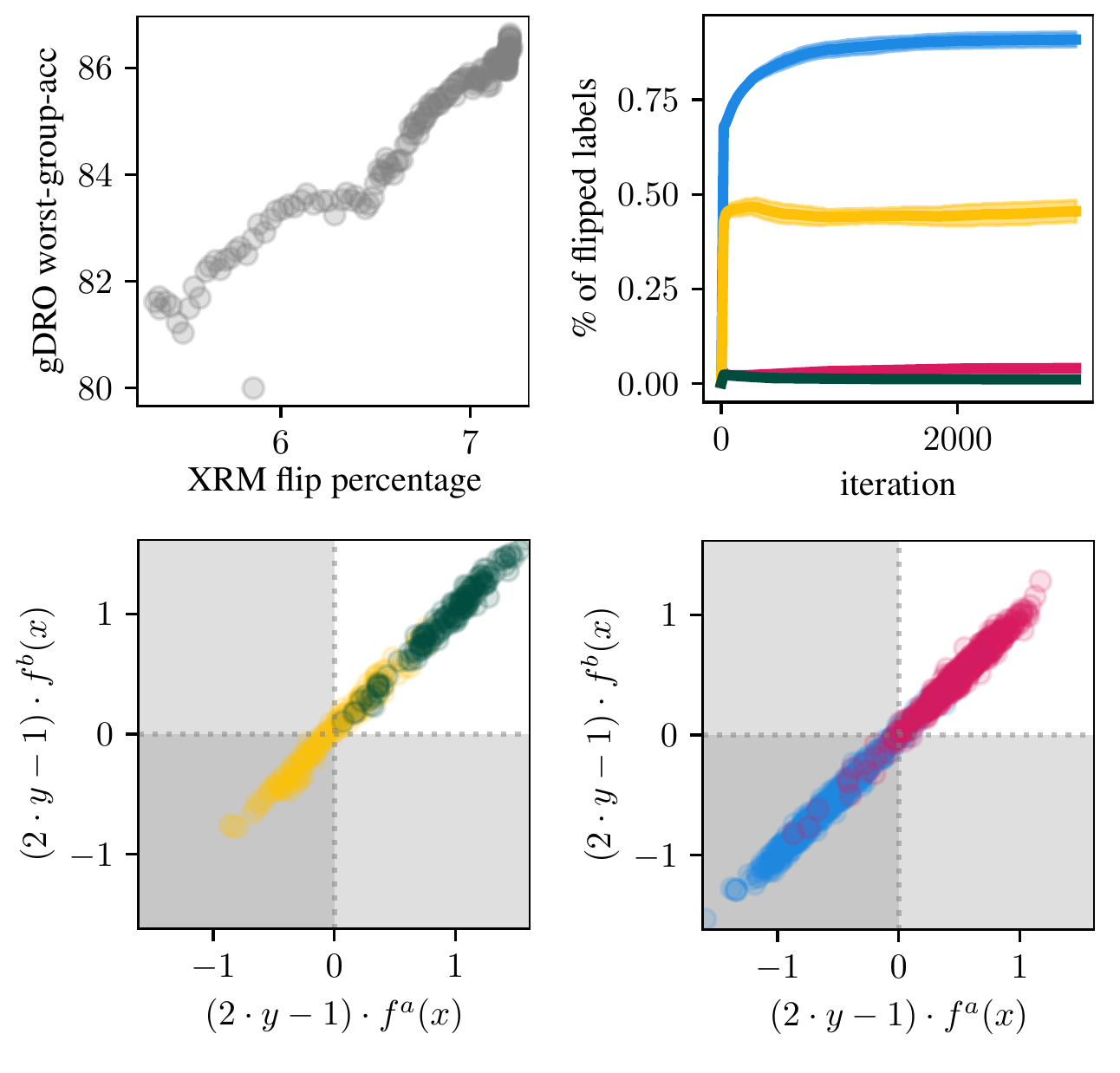}
\definecolor{ww}{HTML}{D81B60}
\definecolor{wl}{HTML}{1E88E5}
\definecolor{lw}{HTML}{FFC107}
\definecolor{ll}{HTML}{004D40}
\vspace{-0.8cm}
\caption{\method{} on the Waterbirds problem, concerning \sqboxblack{ww} waterbirds in water, \sqboxblack{wl} waterbirds in land, \sqboxblack{lw} landbirds in water, \sqboxblack{ll} landbirds in land.
The top-left panel shows that ``percentage of \method{} label-flips at convergence'' is a strong indicator of ``worst-group-accuracy of OOD generalization algorithm'', making flips a good criterion to select twin hyper-parameters.
The two bottom panels show the signed margin of the twins on each ground-truth group. Each of the bottom plots correspond to one of the classes. Note that a positive margin means correct classification.
From each of these class-dependent plots, \method{} discovers two environments: one for points in the ``mistake-free'' white area, and one for points in the ``cross-mistake'' gray areas.
Notably, \method{} is able to allocate the two smallest groups \sqboxblack{wl}\sqboxblack{lw} to dedicated environments.
Another notable observation is that the two bottom plots appear as straight lines, indicating that the twin networks agree on their predictions.
The top-right panel shows that label flipping happens almost exclusively for the two smallest groups, and stabilizes as training progresses.
}
\label{fig:wb}
\end{figure}

\begin{figure}[t!]
    \centering
    \includegraphics[width=0.42\textwidth]{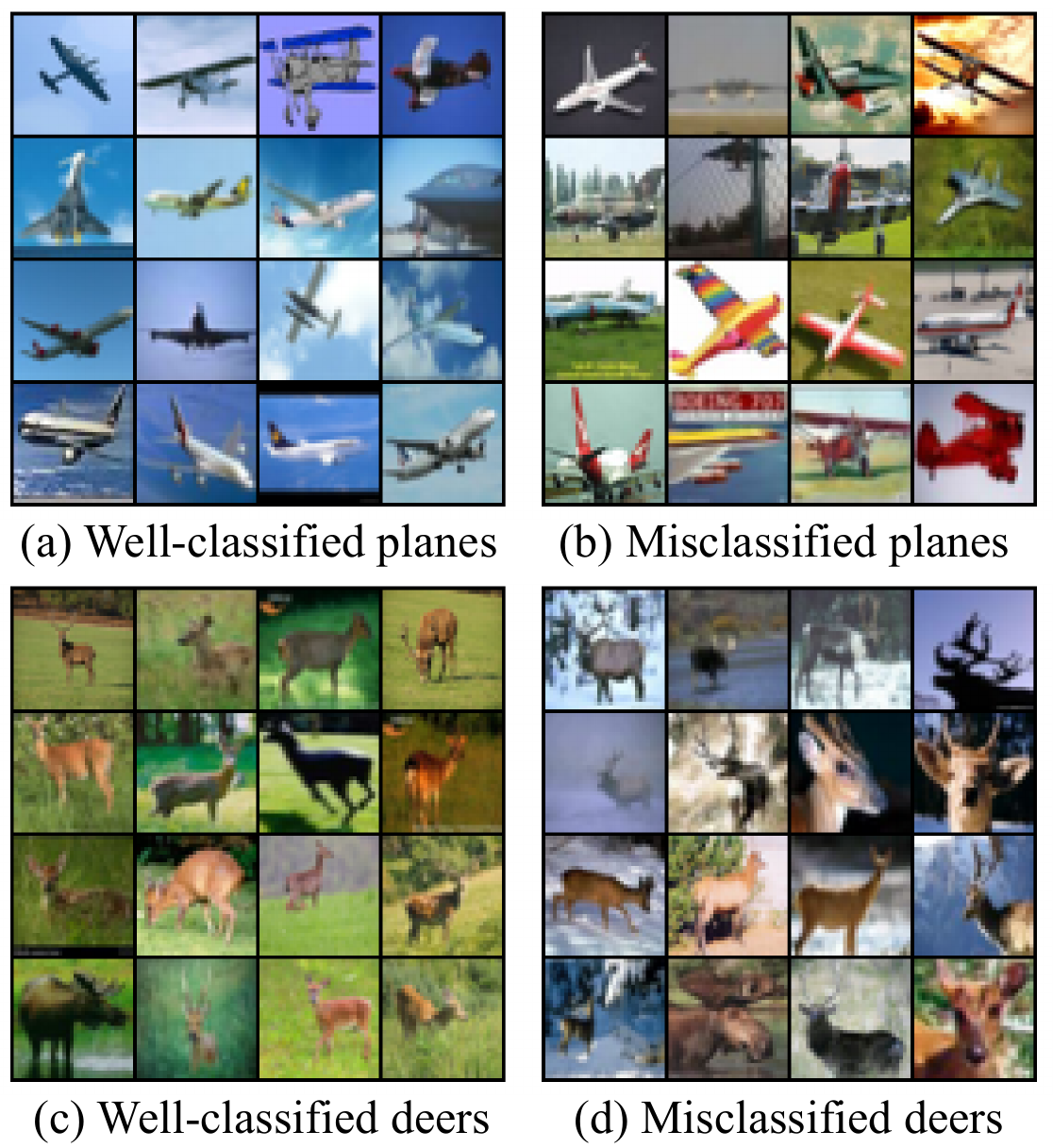}
    \caption{
    Randomly selected images from CIFAR-10, identified by \method{}.
    Although CIFAR-10 lacks predefined environment annotations, our method has successfully uncovered intriguing environments.
    Notably, well-classified examples (when held-out) are prototypical, featuring planes in blue skies and deer on green landscapes.
    In contrast, misclassified examples (when held-out) are less typical, which means they are correctly classified only when included in the training set.
    }
    \vspace{-0.8cm}
    \label{fig:cifar}
\end{figure}

\paragraph{Label flipping dynamics on Waterbirds}
\Cref{fig:wb} explores some of the behaviors of \method{} on the Waterbirds dataset.
In particular, the top-left panel justifies the use of ``percentage of label-flips at convergence'' as a model selection criterion for \method{}, as it correlates strongly with downstream worst-group-accuracy.
The two bottom panels showcase the clear separation of the minority group ``landbirds/water'' by \method{}, as no landbirds in land are in the cross-mistake area.
The top-right panel shows that label flipping happens almost exclusively for minority groups, and converges alongside \method{} training. 
This provides \method{} with a degree of stability, removing the need for intricate early-stopping criteria. 

\vspace{-0.1cm}
\paragraph{Revealing spurious correlations in CIFAR-10 with XRM}
\Cref{fig:cifar} applies \method{} to the CIFAR-10 dataset~\citep{cifar}.
While CIFAR-10 does not contain environment annotations, the discovered environments by \method{} for the ``plane'' and ``deer'' classes reveal one interesting spurious correlation, namely the background color.

As a final remark, we ablated the need for (i) holding-out data, and (ii) performing label flipping, finding that both components are essential to the performance of \method{}.

\section{When Does XRM Fail?}
\label{sec:fail}
In the previous section, we showcased the effectiveness of XRM in successfully discovering relevant environments.
However, it is important to note that XRM, like other environment discovery methods, relies on certain assumptions.
In the absence of these assumptions, these methods, including XRM, may fail to discover relevant environments.
In its full generality, the problem of learning invariant predictors in the absence of appropriate environment annotations is indeed impossible~\citep{lin2022zin}.
To see that, we note that dividing data into invariance-affording environments parallels the problem of discovering the right causal structure in the field of causal inference~\citep{pearl2009causality}.
To reveal the true causal structure between an invariant feature $X_\text{inv}$ and a target $Y$, one must ``control-for'' a set of variables/environments $E$ that satisfy the following \textit{conditional independence statement},
\begin{equation}
Y \perp E \mid X_\text{inv}.
\label{eq:ci}
\end{equation}
Unfortunately, different causal structures can produce identical observational data~\cite{peters2017elements}.
Because of that, identifying an appropriate $E$ is impossible without admitting extra knowledge about the causal structure behind our data, and XRM is not an exception of such free lunch.

Therefore, we would expect XRM to work well in instances where the inferred environments $E$ satisfy~\eqref{eq:ci}, and we should anticipate trouble in those cases where the discovered environments violate~\eqref{eq:ci}.
On the other hand, as discussed in~\citep{lin2022zin}, evaluating~\eqref{eq:ci} requires knowing the invariant feature $X_\text{inv}$, which is the variable subject to discovery.
This makes~\eqref{eq:ci} difficult to verify in practice when inferring environments $E$, and the best we can do is to offer some canonical examples of successes and failures, that can guide our choices of when to apply XRM.

Here, we exemplify with four different versions of the ColorMNIST dataset~\citep{irm}.
All four versions instantiate a colored digit classification task, differing on whether the invariant feature is ``digit shape'' or ``digit color'', and which one of these two variables bear the strongest correlation to the target label.
Overall, we expect ``digit color'' to be faster (easier) to learn, leading to generalization issues when ``digit shape''---more difficult and slower to learn---is the desired invariant feature.

\newcommand\triplet[5]{
  \node[inner sep=0pt] (A) at (#1+0, #2) {$S$};
  \node[inner sep=0pt] (B) at (#1+0.8, #2) {$Y$};
  \node[inner sep=0pt] (C) at (#1+1.6, #2) {$C$};
  \ifnum0=#5\relax
    \draw[->, thick, shorten <=2pt, shorten >=2pt] (A) -- node[above,] {\scriptsize #3} (B);
    \draw[->, thick, shorten <=2pt, shorten >=2pt] (B) -- node[above,] {\scriptsize #4} (C);
  \else
    \draw[->, thick, shorten <=2pt, shorten >=2pt] (C) -- node[above,] {\scriptsize #4} (B);
    \draw[->, thick, shorten <=2pt, shorten >=2pt] (B) -- node[above,] {\scriptsize #3} (A);
  \fi
}

\begin{table}
\resizebox{0.5\textwidth}{!}{%
\begin{tikzpicture}
  \draw[very thick] (-2.75, 1.6) -- (8.3, 1.6);
  
  \node[] (L1) at (0, 1.3) {CMNIST};
  \node[] (L1) at (0, 0.85) {\scriptsize \cite{irm}};
  \node[] (L1) at (2.5, 1.3) {InvCMNIST};
  \node[] (L1) at (2.5, 0.85) {\scriptsize \cite{leon}};
  \node[] (L1) at (5, 1.3) {InveCOLOR};
  \node[] (L1) at (5, 0.85) {\scriptsize (No citation)};
  \node[] (L1) at (7.5, 1.3) {MCOLOR};
  \node[] (L1) at (7.5, 0.85) {\scriptsize \cite{zin}};

  \draw[thick] (-2.75, 0.6) -- (8.3, 0.6);
  \node[text width=1.2cm] (R1) at (-2.2, -0) {training\\($e=1$)};
  \node[text width=1.2cm] (R2) at (-2.2, -1) {training\\($e=2$)}; 
  \node[text width=1.2cm] (R3) at (-2.2, -2) {training\\(pooled)};
  \node[text width=1.2cm] (R4) at (-2.2, -3) {testing};

  \draw[draw=green!15, fill=green!15] (-1.2, -1.5) rectangle ++(2.1, -0.85);
  \draw[draw=yellow!15, fill=yellow!15] (1.3, -1.5) rectangle ++(2.1, -0.85);
  \draw[draw=yellow!15, fill=yellow!15] (3.8, -1.5) rectangle ++(2.1, -0.85);
  \draw[draw=green!15, fill=green!15] (6.3, -1.5) rectangle ++(2.1, -0.85);
  
  \triplet{0-1}{-0}{0.75}{0.80}{0};
  \triplet{0-1}{-1}{0.75}{0.90}{0};
  \triplet{0-1}{-2}{0.75}{0.85}{0};
  \triplet{0-1}{-3}{0.75}{0.10}{0};
  \triplet{2.5-1}{-0}{0.85}{0.70}{0};
  \triplet{2.5-1}{-1}{0.85}{0.80}{0};
  \triplet{2.5-1}{-2}{0.85}{0.75}{0};
  \triplet{2.5-1}{-3}{0.85}{0.10}{0};
  \triplet{5-1}{-0}{0.80}{0.75}{1};
  \triplet{5-1}{-1}{0.90}{0.75}{1};
  \triplet{5-1}{-2}{0.85}{0.75}{1};
  \triplet{5-1}{-3}{0.10}{0.75}{1};
  \triplet{7.5-1}{-0}{0.80}{0.85}{1};
  \triplet{7.5-1}{-1}{0.70}{0.85}{1};
  \triplet{7.5-1}{-2}{0.75}{0.85}{1};
  \triplet{7.5-1}{-3}{0.10}{0.85}{1};

  \draw[thick] (-2.75, -3.5) -- (8.3, -3.5);

  \node[text width=1.2cm] (R4) at (-2.2, -4+0.1) {inv.feat.};
  \node[] (X) at (0, -4+0.1) {complex, weak};
  \node[] (X) at (2.5, -4+0.1) {complex, strong};
  \node[] (X) at (5, -4+0.1) {simple, weak};
  \node[] (X) at (7.5, -4+0.1) {simple, strong};
  
  \node[text width=1.2cm] (R4) at (-2.2, -4.5) {ERM};
  \node[] (X) at (0, -4.5) {$0.37 \pm 0.10$};
  \node[] (X) at (2.5, -4.5) {$0.67 \pm 0.02$};
  \node[] (X) at (5, -4.5) {$\textbf{0.75} \pm 0.01$};
  \node[] (X) at (7.5, -4.5) {$\textbf{0.85} \pm 0.01$};
  
  \node[text width=1.2cm] (R4) at (-2.2, -5) {XRM};
  \node[] (X) at (0, -5) {$\textbf{0.71} \pm 0.02$};
  \node[] (X) at (2.5, -5) {$\textbf{0.82} \pm 0.02$};
  \node[] (X) at (5, -5) {$0.40 \pm 0.01$};
  \node[] (X) at (7.5, -5) {$0.57 \pm 0.01$};
  
  \node[text width=1.2cm] (R4) at (-2.2, -5.5) {Oracle};
  \node[] (X) at (0, -5.5) {$0.75$};
  \node[] (X) at (2.5, -5.5) {$0.85$};
  \node[] (X) at (5, -5.5) {$0.75$};
  \node[] (X) at (7.5, -5.5) {$0.85$};

  \draw[very thick] (-2.75, -6) -- (8.3, -6);
\end{tikzpicture}}
\caption{%
Four ColoredMNIST versions, where the environment $E$ influences digit shape $S$ and color $C$, forming our input $X = (S, C)$.
We depict the causal structure for each dataset version, and the correlation between variables.
The invariant feature may be the complex digit shape (CMNIST versions) \emph{or} the simple digit color (MCOLOR versions), which in turn could bear the strongest \emph{or} weakest correlation to the target variable---producing four versions of the ColoredMNIST problem.
Note that \colorbox{green!15}{CMNIST}-\colorbox{green!15}{MCOLOR} and \colorbox{yellow!15}{InverseCMNIST}-\colorbox{yellow!15}{InverseMCOLOR} are indistinguishable from pooled training data alone. 
At the bottom, test accuracies of ERM, XRM+GroupDRO, and an Oracle which relies solely on the invariant feature.}
\vspace{-0.5cm}
\label{table:fourcolors}
\end{table}

We show in~\cref{table:fourcolors} the average-test-accuracy of ERM and XRM followed by GroupDRO for the four versions of the ColoredMNIST dataset. We also show what a hypothetical oracle, relying solely on the invariant feature, would achieve.
ERM performs well when the invariant feature is the simplest of the two.
XRM performs well when the invariant feature is the most complex of the two.
We highlight that the datasets CMNIST and MCOLOR are observationally equivalent from pooled data alone---and a similar remark follows for InverseCMNIST and InverseMCOLOR.
This echoes the impossibility results of~\citep{lin2022zin}, namely learning invariant predictors in the absence of environment annotations is impossible in its full generality: for instance, based on training data alone, we would never know if we are dealing with InverseCMNIST or InverseMCOLOR, and therefore we are at a loss of whether to apply ERM or XRM.
Nevertheless, XRM remains an state-of-the-art solution for those problems were we would like our learning machine to ignore the fastest-to-learn feature, often being a spurious shortcut~\citep{geirhos2020shortcut,shah2020pitfalls,gs}, in order to focus on more complex patterns with a higher potential for invariance.

\section{Discussion}
\label{sec:discussion}

We have introduced \fullmethod{} (\method{}), a simple algorithm for environment discovery.
\method{} provides a recipe to tune its hyper-parameters, does not require early-stopping, and can discover environments for all training and validation data---dropping the requirement for human annotations at all.
More specifically, \method{} trains two twin classifiers on random halves of the training data, while encouraging each twin to imitate confident held-out mistakes by their sibling.
This implements an ``echo-chamber'' that identifies environments that differ only in spurious correlation, and endow subsequent OOD generalization algorithms with oracle-like performance.

We highlight two directions for future work.
Firstly, how does \method{} relate to the invariance principle $Y \perp E \mid \Phi(X)$?
What is the interplay between revealing relevant labels $Y$ and relevant environments $E$ as to afford invariance?
To our knowledge, \method{} is the first environment discovery algorithm tampering with labels $Y$, thus exploring invariance---and the violation thereof---from a new angle.
Because relabeling happens with a probability proportional to confidence, we expect model calibration to play a role in understanding the theoretical underpinnings of \method{}, as it happened with other invariance methods~\citep{dg_calib}.
Overall, the theoretical analysis of \method{} will call for new tools, because label-flipping steers XRM away from the Bayes-optimal predictor.

Secondly, we would like to further understand the relationship between \method{} and the multifarious phenomenon of memorization.
Good memorization affords invariance (\emph{Where did I park my car?}), and therefore depends on the collection of environments deemed relevant.
Bad memorization happens due to ``structured over-fitting'', commonly incarnated as a bad learning strategy; ``use a simple feature for the majority, then memorize the minority''.
\method{} seems to attack a similar problem but, how does it specifically relate to these two flavours of memorization?
Does \method{} discover environments that promote features that benefit all examples?

\section*{Acknowledgements}
We thank Kartik Ahuja, Mart\'in Arjovsky, L\'eon Bottou, Kamalika Chaudhuri, Cijo Jose, Reyhane Askari Hemmat and Polina Kirichenko for fruitful discussions. We thank the FAIR leadership for their help in publishing this manuscript.

\section*{Impact Statement}
This paper presents work whose goal is to advance the field of Machine Learning. There are many potential societal consequences of our work, none which we feel must be specifically highlighted here.

\nocite{langley00}
\bibliography{example_paper}

\begin{thebibliography}{84}
\providecommand{\natexlab}[1]{#1}
\providecommand{\url}[1]{\texttt{#1}}
\expandafter\ifx\csname urlstyle\endcsname\relax
  \providecommand{\doi}[1]{doi: #1}\else
  \providecommand{\doi}{doi: \begingroup \urlstyle{rm}\Url}\fi

\bibitem[Angwin et~al.(2016)Angwin, Larson, Mattu, and Kirchner]{ai_justice}
Angwin, J., Larson, J., Mattu, S., and Kirchner, L.
\newblock Machine bias.
\newblock \emph{ProPublica}, 2016.
\newblock URL
  \url{https://www.propublica.org/article/machine-bias-risk-assessments\\-in-criminal-sentencing}.

\bibitem[Arjovsky et~al.(2019)Arjovsky, Bottou, Gulrajani, and Lopez-Paz]{irm}
Arjovsky, M., Bottou, L., Gulrajani, I., and Lopez-Paz, D.
\newblock Invariant risk minimization.
\newblock \emph{arXiv}, 2019.
\newblock URL \url{https://arxiv.org/abs/1907.02893}.

\bibitem[{Bao} \& {Barzilay}(2022){Bao} and {Barzilay}]{ls}
{Bao}, Y. and {Barzilay}, R.
\newblock Learning to split for automatic bias detection.
\newblock \emph{arXiv}, 2022.
\newblock URL \url{https://arxiv.org/abs/2204.13749}.

\bibitem[Barocas et~al.(2019)Barocas, Hardt, and
  Narayanan]{barocas-hardt-narayanan}
Barocas, S., Hardt, M., and Narayanan, A.
\newblock \emph{Fairness and Machine Learning: Limitations and Opportunities}.
\newblock fairmlbook.org, 2019.
\newblock \url{http://www.fairmlbook.org}.

\bibitem[Bell \& Sagun(2023)Bell and Sagun]{bell2023simplicity}
Bell, S.~J. and Sagun, L.
\newblock Simplicity bias leads to amplified performance disparities.
\newblock In \emph{Proceedings of the 2023 ACM Conference on Fairness,
  Accountability, and Transparency}, pp.\  355--369, 2023.

\bibitem[Blanchard et~al.(2011)Blanchard, Lee, and Scott]{blanchard}
Blanchard, G., Lee, G., and Scott, C.
\newblock Generalizing from several related classification tasks to a new
  unlabeled sample.
\newblock In \emph{NeurIPS}, 2011.
\newblock URL
  \url{https://proceedings.neurips.cc/paper_files/paper/2011/file/b571ecea16a9824023ee1af16897a582-Paper.pdf}.

\bibitem[Bojarski et~al.(2016)Bojarski, Del~Testa, Dworakowski, Firner, Flepp,
  Goyal, Jackel, Monfort, Muller, Zhang, et~al.]{ai_driving}
Bojarski, M., Del~Testa, D., Dworakowski, D., Firner, B., Flepp, B., Goyal, P.,
  Jackel, L.~D., Monfort, M., Muller, U., Zhang, J., et~al.
\newblock End to end learning for self-driving cars.
\newblock \emph{arXiv}, 2016.
\newblock URL \url{https://arxiv.org/abs/1604.07316}.

\bibitem[Borkan et~al.(2019)Borkan, Dixon, Sorensen, Thain, and
  Vasserman]{borkan2019nuanced}
Borkan, D., Dixon, L., Sorensen, J., Thain, N., and Vasserman, L.
\newblock Nuanced metrics for measuring unintended bias with real data for text
  classification.
\newblock \emph{WWW}, 2019.
\newblock URL \url{https://arxiv.org/abs/1903.04561}.

\bibitem[Bottou(2019)]{bottou_shuffle}
Bottou, L.
\newblock Learning representation with causal invariance.
\newblock \emph{ICLR Keynote}, 2019.
\newblock URL \url{https://leon.bottou.org/talks/invariances}.

\bibitem[Cha et~al.(2021)Cha, Chun, Lee, Cho, Park, Lee, and Park]{cha2021swad}
Cha, J., Chun, S., Lee, K., Cho, H.-C., Park, S., Lee, Y., and Park, S.
\newblock {SWAD}: Domain generalization by seeking flat minima.
\newblock \emph{NeurIPS}, 2021.
\newblock URL \url{https://arxiv.org/abs/2102.08604}.

\bibitem[Chen et~al.(2022)Chen, Zhou, Bian, Xie, Wu, Zhang, Ma, Yang, Zhao,
  Han, et~al.]{chen2022pareto}
Chen, Y., Zhou, K., Bian, Y., Xie, B., Wu, B., Zhang, Y., Ma, K., Yang, H.,
  Zhao, P., Han, B., et~al.
\newblock Pareto invariant risk minimization: Towards mitigating the
  optimization dilemma in out-of-distribution generalization.
\newblock \emph{arXiv preprint arXiv:2206.07766}, 2022.

\bibitem[Chen et~al.(2023)Chen, Bian, Zhou, Xie, Han, and
  Cheng]{chen2023rethinking}
Chen, Y., Bian, Y., Zhou, K., Xie, B., Han, B., and Cheng, J.
\newblock Rethinking invariant graph representation learning without
  environment partitions.
\newblock \emph{ICLR DG Workshop}, 2023.
\newblock URL \url{https://openreview.net/forum?id=bjw5jqGtDy}.

\bibitem[Chen et~al.(2024)Chen, Bian, Zhou, Xie, Han, and Cheng]{chen2024does}
Chen, Y., Bian, Y., Zhou, K., Xie, B., Han, B., and Cheng, J.
\newblock Does invariant graph learning via environment augmentation learn
  invariance?
\newblock \emph{Advances in Neural Information Processing Systems}, 36, 2024.

\bibitem[{Creager} et~al.(2020){Creager}, {Jacobsen}, and {Zemel}]{eiil}
{Creager}, E., {Jacobsen}, J.-H., and {Zemel}, R.
\newblock Environment inference for invariant learning.
\newblock \emph{arXiv}, 2020.
\newblock URL \url{https://arxiv.org/abs/2010.07249}.

\bibitem[{Dagaev} et~al.(2021){Dagaev}, {Roads}, {Luo}, {Barry}, {Patil}, and
  {Love}]{toogood}
{Dagaev}, N., {Roads}, B.~D., {Luo}, X., {Barry}, D.~N., {Patil}, K.~R., and
  {Love}, B.~C.
\newblock A too-good-to-be-true prior to reduce shortcut reliance.
\newblock \emph{arXiv}, 2021.
\newblock URL \url{https://arxiv.org/abs/2102.06406}.

\bibitem[DeGrave et~al.(2021)DeGrave, Janizek, and Lee]{covid}
DeGrave, A.~J., Janizek, J.~D., and Lee, S.-I.
\newblock {AI} for radiographic {COVID-19} detection selects shortcuts over
  signal.
\newblock \emph{Nature Machine Intelligence}, 2021.
\newblock URL \url{https://www.nature.com/articles/s42256-021-00338-7}.

\bibitem[Devlin et~al.(2018)Devlin, Chang, Lee, and Toutanova]{devlin2018bert}
Devlin, J., Chang, M.-W., Lee, K., and Toutanova, K.
\newblock Bert: Pre-training of deep bidirectional transformers for language
  understanding.
\newblock \emph{arXiv preprint arXiv:1810.04805}, 2018.

\bibitem[Eastwood et~al.(2023)Eastwood, Singh, Nicolicioiu, Vlastelica, von
  K{\"u}gelgen, and Sch{\"o}lkopf]{eastwood2023spuriosity}
Eastwood, C., Singh, S., Nicolicioiu, A.~L., Vlastelica, M., von K{\"u}gelgen,
  J., and Sch{\"o}lkopf, B.
\newblock Spuriosity didn't kill the classifier: Using invariant predictions to
  harness spurious features.
\newblock \emph{arXiv}, 2023.
\newblock URL \url{https://arxiv.org/abs/2307.09933}.

\bibitem[{Feldman} \& {Zhang}(2020){Feldman} and {Zhang}]{feldman}
{Feldman}, V. and {Zhang}, C.
\newblock What neural networks memorize and why: Discovering the long tail via
  influence estimation.
\newblock \emph{arXiv}, 2020.
\newblock URL \url{https://arxiv.org/abs/2008.03703}.

\bibitem[Ganin et~al.(2016)Ganin, Ustinova, Ajakan, Germain, Larochelle,
  Laviolette, Marchand, and Lempitsky]{dann}
Ganin, Y., Ustinova, E., Ajakan, H., Germain, P., Larochelle, H., Laviolette,
  F., Marchand, M., and Lempitsky, V.
\newblock Domain-adversarial training of neural networks.
\newblock \emph{JMLR}, 2016.
\newblock URL \url{https://arxiv.org/abs/1505.07818}.

\bibitem[Geirhos et~al.(2020)Geirhos, Jacobsen, Michaelis, Zemel, Brendel,
  Bethge, and Wichmann]{geirhos2020shortcut}
Geirhos, R., Jacobsen, J.-H., Michaelis, C., Zemel, R., Brendel, W., Bethge,
  M., and Wichmann, F.~A.
\newblock Shortcut learning in deep neural networks.
\newblock \emph{Nature Machine Intelligence}, 2020.
\newblock URL \url{https://arxiv.org/abs/2004.07780}.

\bibitem[Giannone et~al.(2022)Giannone, Havrylov, Massiah, Yilmaz, and
  Jiao]{giannone2022just}
Giannone, G., Havrylov, S., Massiah, J., Yilmaz, E., and Jiao, Y.
\newblock Just mix once: Worst-group generalization by group interpolation.
\newblock \emph{NeurIPS Workshop on Distribution Shifts}, 2022.
\newblock URL \url{https://arxiv.org/abs/2210.12195}.

\bibitem[Goodfellow et~al.(2014)Goodfellow, Shlens, and Szegedy]{adversarial}
Goodfellow, I.~J., Shlens, J., and Szegedy, C.
\newblock Explaining and harnessing adversarial examples.
\newblock \emph{arXiv}, 2014.
\newblock URL \url{https://arxiv.org/abs/1412.6572}.

\bibitem[Gulrajani \& Lopez-Paz(2020)Gulrajani and Lopez-Paz]{domainbed}
Gulrajani, I. and Lopez-Paz, D.
\newblock In search of lost domain generalization.
\newblock \emph{arXiv}, 2020.
\newblock URL \url{https://arxiv.org/abs/2007.01434}.

\bibitem[Hand \& Henley(1997)Hand and Henley]{ai_finance}
Hand, D.~J. and Henley, W.~E.
\newblock Statistical classification methods in consumer credit scoring: a
  review.
\newblock \emph{Journal of the royal statistical society: series a (statistics
  in society)}, 1997.
\newblock URL \url{https://www.jstor.org/stable/2983268}.

\bibitem[He et~al.(2016)He, Zhang, Ren, and Sun]{he2016deep}
He, K., Zhang, X., Ren, S., and Sun, J.
\newblock Deep residual learning for image recognition.
\newblock In \emph{Proceedings of the IEEE conference on computer vision and
  pattern recognition}, pp.\  770--778, 2016.

\bibitem[Heaven(2021)]{covid2}
Heaven, W.~D.
\newblock Hundreds of {AI} tools have been built to catch {COVID}. none of them
  helped.
\newblock \emph{MIT Technology Review}, 2021.
\newblock URL
  \url{https://www.technologyreview.com/2021/07/30/1030329/machine-learning-ai-failed//-covid-hospital-diagnosis-pandemic/}.

\bibitem[Idrissi et~al.(2022)Idrissi, Arjovsky, Pezeshki, and Lopez-Paz]{subg}
Idrissi, B.~Y., Arjovsky, M., Pezeshki, M., and Lopez-Paz, D.
\newblock Simple data balancing achieves competitive worst-group-accuracy.
\newblock \emph{CLeaR}, 2022.
\newblock URL \url{https://arxiv.org/abs/2110.14503}.

\bibitem[Izmailov et~al.(2022)Izmailov, Kirichenko, Gruver, and
  Wilson]{izmailov2022feature}
Izmailov, P., Kirichenko, P., Gruver, N., and Wilson, A.~G.
\newblock On feature learning in the presence of spurious correlations.
\newblock \emph{NeurIPS}, 2022.
\newblock URL \url{https://arxiv.org/abs/2210.11369}.

\bibitem[Japkowicz(2000)]{japkowicz2000class}
Japkowicz, N.
\newblock The class imbalance problem: Significance and strategies.
\newblock In \emph{ICML}, 2000.
\newblock URL
  \url{https://www.researchgate.net/publication/2639031_The_Class_Imbalance_Problem_Significance_and_Strategies}.

\bibitem[Jiang et~al.(2017)Jiang, Jiang, Zhi, Dong, Li, Ma, Wang, Dong, Shen,
  and Wang]{ai_healthcare}
Jiang, F., Jiang, Y., Zhi, H., Dong, Y., Li, H., Ma, S., Wang, Y., Dong, Q.,
  Shen, H., and Wang, Y.
\newblock Artificial intelligence in healthcare: past, present and future.
\newblock \emph{Stroke and vascular neurology}, 2017.
\newblock URL \url{https://pubmed.ncbi.nlm.nih.gov/29507784/}.

\bibitem[Kim et~al.(2023)Kim, Baratin, Zhang, and
  Lacoste-Julien]{kim2023crosssplit}
Kim, J., Baratin, A., Zhang, Y., and Lacoste-Julien, S.
\newblock Crosssplit: mitigating label noise memorization through data
  splitting.
\newblock In \emph{International Conference on Machine Learning}, pp.\
  16377--16392. PMLR, 2023.

\bibitem[Kirichenko et~al.(2022)Kirichenko, Izmailov, and
  Wilson]{kirichenko2022last}
Kirichenko, P., Izmailov, P., and Wilson, A.~G.
\newblock Last layer re-training is sufficient for robustness to spurious
  correlations.
\newblock \emph{arXiv preprint arXiv:2204.02937}, 2022.

\bibitem[Krizhevsky et~al.(2009)Krizhevsky, Nair, and Hinton]{cifar}
Krizhevsky, A., Nair, V., and Hinton, G.
\newblock Cifar-10, 2009.
\newblock URL \url{http://www.cs.toronto.edu/~kriz/cifar.html}.

\bibitem[Krueger et~al.(2021)Krueger, Caballero, Jacobsen, Zhang, Binas, Zhang,
  Le~Priol, and Courville]{krueger2021out}
Krueger, D., Caballero, E., Jacobsen, J.-H., Zhang, A., Binas, J., Zhang, D.,
  Le~Priol, R., and Courville, A.
\newblock Out-of-distribution generalization via risk extrapolation (rex).
\newblock In \emph{ICML}, 2021.
\newblock URL \url{https://arxiv.org/abs/2003.00688}.

\bibitem[LaBonte et~al.(2023)LaBonte, Muthukumar, and
  Kumar]{labonte2023towards}
LaBonte, T., Muthukumar, V., and Kumar, A.
\newblock Towards last-layer retraining for group robustness with fewer
  annotations.
\newblock \emph{arXiv preprint arXiv:2309.08534}, 2023.

\bibitem[{Lahoti} et~al.(2020){Lahoti}, {Beutel}, {Chen}, {Lee}, {Prost},
  {Thain}, {Wang}, and {Chi}]{arl}
{Lahoti}, P., {Beutel}, A., {Chen}, J., {Lee}, K., {Prost}, F., {Thain}, N.,
  {Wang}, X., and {Chi}, E.~H.
\newblock Fairness without demographics through adversarially reweighted
  learning.
\newblock \emph{arXiv}, 2020.
\newblock URL \url{https://arxiv.org/abs/2006.13114}.

\bibitem[Lee et~al.(2023)Lee, Yao, and Finn]{lee2022diversify}
Lee, Y., Yao, H., and Finn, C.
\newblock Diversify and disambiguate: Learning from underspecified data.
\newblock \emph{ICLR}, 2023.
\newblock URL \url{https://arxiv.org/abs/2202.03418}.

\bibitem[Li et~al.(2023)Li, Liu, and Hu]{bam}
Li, G., Liu, J., and Hu, W.
\newblock Bias amplification enhances minority group performance.
\newblock \emph{arXiv}, 2023.
\newblock URL \url{https://arxiv.org/abs/2309.06717}.

\bibitem[Liang \& Zou(2022)Liang and Zou]{liang2022metashift}
Liang, W. and Zou, J.
\newblock Metashift: A dataset of datasets for evaluating contextual
  distribution shifts and training conflicts.
\newblock \emph{arXiv}, 2022.
\newblock URL \url{https://arxiv.org/abs/2202.06523}.

\bibitem[Lin et~al.(2022{\natexlab{a}})Lin, Zhu, Tan, and Cui]{lin2022zin}
Lin, Y., Zhu, S., Tan, L., and Cui, P.
\newblock Zin: When and how to learn invariance without environment partition?
\newblock \emph{NeurIPS}, 2022{\natexlab{a}}.
\newblock URL \url{https://arxiv.org/abs/2203.05818}.

\bibitem[Lin et~al.(2022{\natexlab{b}})Lin, Zhu, Tan, and Cui]{zin}
Lin, Y., Zhu, S., Tan, L., and Cui, P.
\newblock Zin: When and how to learn invariance without environment partition?
\newblock \emph{Advances in Neural Information Processing Systems},
  35:\penalty0 24529--24542, 2022{\natexlab{b}}.

\bibitem[Lin et~al.(2023)Lin, Tan, Hao, Wong, Dong, Zhang, Yang, and
  Zhang]{lin2023spurious}
Lin, Y., Tan, L., Hao, Y., Wong, H., Dong, H., Zhang, W., Yang, Y., and Zhang,
  T.
\newblock Spurious feature diversification improves out-of-distribution
  generalization.
\newblock \emph{arXiv preprint arXiv:2309.17230}, 2023.
\newblock URL \url{https://arxiv.org/abs/2309.17230}.

\bibitem[Liu et~al.(2015)Liu, Luo, Wang, and Tang]{celeba}
Liu, Z., Luo, P., Wang, X., and Tang, X.
\newblock Deep learning face attributes in the wild.
\newblock In \emph{ICCV}, 2015.
\newblock URL \url{https://arxiv.org/abs/1411.7766}.

\bibitem[Lopez-Paz et~al.(2022)Lopez-Paz, Bouchacourt, Sagun, and
  Usunier]{lopez2022measuring}
Lopez-Paz, D., Bouchacourt, D., Sagun, L., and Usunier, N.
\newblock Measuring and signing fairness as performance under multiple
  stakeholder distributions.
\newblock \emph{arXiv}, 2022.
\newblock URL \url{https://arxiv.org/abs/2207.09960}.

\bibitem[Loshchilov \& Hutter(2017)Loshchilov and
  Hutter]{loshchilov2017decoupled}
Loshchilov, I. and Hutter, F.
\newblock Decoupled weight decay regularization.
\newblock \emph{arXiv preprint arXiv:1711.05101}, 2017.

\bibitem[Ma et~al.(2023)Ma, Yue, Tomoyuki, Tomoki, Jayashree, Pranata, and
  Zhang]{ma2023invariant}
Ma, J., Yue, Z., Tomoyuki, K., Tomoki, S., Jayashree, K., Pranata, S., and
  Zhang, H.
\newblock Invariant feature regularization for fair face recognition.
\newblock In \emph{ICCV}, 2023.
\newblock URL
  \url{https://openaccess.thecvf.com/content/ICCV2023/papers/Ma_Invariant_Feature_Regularization_for_Fair_Face_Recognition_ICCV_2023_paper.pdf}.

\bibitem[Muandet et~al.(2013)Muandet, Balduzzi, and
  Schölkopf]{pmlr-v28-muandet13}
Muandet, K., Balduzzi, D., and Schölkopf, B.
\newblock Domain generalization via invariant feature representation.
\newblock In \emph{ICML}, 2013.
\newblock URL \url{https://proceedings.mlr.press/v28/muandet13.html}.

\bibitem[{Nam} et~al.(2020){Nam}, {Cha}, {Ahn}, {Lee}, and {Shin}]{lff}
{Nam}, J., {Cha}, H., {Ahn}, S., {Lee}, J., and {Shin}, J.
\newblock Learning from failure: Training debiased classifier from biased
  classifier.
\newblock \emph{arXiv}, 2020.
\newblock URL \url{https://arxiv.org/abs/2007.02561}.

\bibitem[Pagliardini et~al.(2023)Pagliardini, Jaggi, Fleuret, and
  Karimireddy]{pagliardini2022agree}
Pagliardini, M., Jaggi, M., Fleuret, F., and Karimireddy, S.~P.
\newblock Agree to disagree: Diversity through disagreement for better
  transferability.
\newblock \emph{ICLR}, 2023.
\newblock URL \url{https://arxiv.org/abs/2202.04414}.

\bibitem[{Pang Wei Ko et al.}(2021)]{wilds}
{Pang Wei Ko et al.}
\newblock {WILDS}: A benchmark of in-the-wild distribution shifts.
\newblock \emph{arXiv}, 2021.
\newblock URL \url{https://arxiv.org/abs/2012.07421}.

\bibitem[Pearl(2009)]{pearl2009causality}
Pearl, J.
\newblock \emph{Causality}.
\newblock Cambridge university press, 2009.
\newblock URL \url{http://bayes.cs.ucla.edu/BOOK-2K/}.

\bibitem[Peters et~al.(2016)Peters, B{\"u}hlmann, and
  Meinshausen]{peters2016causal}
Peters, J., B{\"u}hlmann, P., and Meinshausen, N.
\newblock Causal inference by using invariant prediction: identification and
  confidence intervals.
\newblock \emph{Journal of the Royal Statistical Society Series B: Statistical
  Methodology}, 2016.
\newblock URL \url{https://arxiv.org/abs/1501.01332}.

\bibitem[Peters et~al.(2017)Peters, Janzing, and
  Sch{\"o}lkopf]{peters2017elements}
Peters, J., Janzing, D., and Sch{\"o}lkopf, B.
\newblock \emph{Elements of causal inference: foundations and learning
  algorithms}.
\newblock The MIT Press, 2017.
\newblock URL
  \url{https://mitpress.mit.edu/9780262037310/elements-of-causal-inference/}.

\bibitem[Pezeshki et~al.(2021)Pezeshki, Kaba, Bengio, Courville, Precup, and
  Lajoie]{gs}
Pezeshki, M., Kaba, O., Bengio, Y., Courville, A.~C., Precup, D., and Lajoie,
  G.
\newblock Gradient starvation: A learning proclivity in neural networks.
\newblock \emph{Advances in Neural Information Processing Systems},
  34:\penalty0 1256--1272, 2021.

\bibitem[Qiu et~al.(2023)Qiu, Potapczynski, Izmailov, and
  Wilson]{qiu2023simple}
Qiu, S., Potapczynski, A., Izmailov, P., and Wilson, A.~G.
\newblock Simple and fast group robustness by automatic feature reweighting.
\newblock \emph{arXiv}, 2023.
\newblock URL \url{https://arxiv.org/abs/2306.11074}.

\bibitem[Rame et~al.(2022)Rame, Kirchmeyer, Rahier, Rakotomamonjy, Gallinari,
  and Cord]{rame2022diverse}
Rame, A., Kirchmeyer, M., Rahier, T., Rakotomamonjy, A., Gallinari, P., and
  Cord, M.
\newblock Diverse weight averaging for out-of-distribution generalization.
\newblock \emph{NeurIPS}, 2022.
\newblock URL \url{https://arxiv.org/abs/2205.09739}.

\bibitem[{Sagawa} et~al.(2019){Sagawa}, {Koh}, {Hashimoto}, and {Liang}]{dro}
{Sagawa}, S., {Koh}, P.~W., {Hashimoto}, T.~B., and {Liang}, P.
\newblock Distributionally robust neural networks for group shifts: On the
  importance of regularization for worst-case generalization.
\newblock \emph{ICLR}, 2019.
\newblock URL \url{https://arxiv.org/abs/1911.08731}.

\bibitem[Shah et~al.(2020{\natexlab{a}})Shah, Tamuly, Raghunathan, Jain, and
  Netrapalli]{shah2020pitfalls}
Shah, H., Tamuly, K., Raghunathan, A., Jain, P., and Netrapalli, P.
\newblock The pitfalls of simplicity bias in neural networks.
\newblock \emph{Advances in Neural Information Processing Systems},
  33:\penalty0 9573--9585, 2020{\natexlab{a}}.

\bibitem[Shah et~al.(2020{\natexlab{b}})Shah, Tamuly, Raghunathan, Jain, and
  Netrapalli]{simplicity}
Shah, H., Tamuly, K., Raghunathan, A., Jain, P., and Netrapalli, P.
\newblock The pitfalls of simplicity bias in neural networks.
\newblock \emph{NeurIPS}, 2020{\natexlab{b}}.
\newblock URL \url{https://arxiv.org/abs/2006.07710}.

\bibitem[Simonyan \& Zisserman(2014)Simonyan and Zisserman]{simonyan2014very}
Simonyan, K. and Zisserman, A.
\newblock Very deep convolutional networks for large-scale image recognition.
\newblock \emph{arXiv preprint arXiv:1409.1556}, 2014.

\bibitem[Sohoni et~al.(2020)Sohoni, Dunnmon, Angus, Gu, and
  R{\'e}]{sohoni2020no}
Sohoni, N., Dunnmon, J., Angus, G., Gu, A., and R{\'e}, C.
\newblock No subclass left behind: Fine-grained robustness in coarse-grained
  classification problems.
\newblock \emph{NeurIPS}, 2020.
\newblock URL \url{https://arxiv.org/abs/2011.12945}.

\bibitem[Sun \& Saenko(2016)Sun and Saenko]{sun2016deep}
Sun, B. and Saenko, K.
\newblock Deep coral: Correlation alignment for deep domain adaptation.
\newblock In \emph{ECCV Workshops}, 2016.

\bibitem[Tan et~al.(2023)Tan, Yong, Zhu, Qu, Qiu, Yinghui, Cui, and
  Qi]{pmlr-v202-tan23b}
Tan, X., Yong, L., Zhu, S., Qu, C., Qiu, X., Yinghui, X., Cui, P., and Qi, Y.
\newblock Provably invariant learning without domain information.
\newblock In \emph{Proceedings of the 40th International Conference on Machine
  Learning}, 2023.
\newblock URL \url{https://proceedings.mlr.press/v202/tan23b.html}.

\bibitem[Teney et~al.(2021)Teney, Abbasnejad, and van~den
  Hengel]{teney2021unshuffling}
Teney, D., Abbasnejad, E., and van~den Hengel, A.
\newblock Unshuffling data for improved generalization in visual question
  answering.
\newblock In \emph{ICCV}, 2021.
\newblock URL \url{https://arxiv.org/abs/2002.11894}.

\bibitem[Tsirigotis et~al.(2023)Tsirigotis, Monteiro, Rodriguez, Vazquez, and
  Courville]{tsirigotis2023group}
Tsirigotis, C., Monteiro, J., Rodriguez, P., Vazquez, D., and Courville, A.
\newblock Group robust classification without any group information.
\newblock \emph{arXiv preprint arXiv:2310.18555}, 2023.

\bibitem[Vapnik(1998)]{vapnik}
Vapnik, V.
\newblock \emph{Statistical learning theory}.
\newblock Wiley, 1998.
\newblock URL
  \url{https://www.wiley.com/en-us/Statistical+Learning+Theory-p-9780471030034}.

\bibitem[Wah et~al.(2011)Wah, Branson, Welinder, Perona, and
  Belongie]{wah2011caltech}
Wah, C., Branson, S., Welinder, P., Perona, P., and Belongie, S.
\newblock The {Caltech-UCSD} birds-200-2011 dataset.
\newblock \emph{California Institute of Technology}, 2011.
\newblock URL \url{https://www.vision.caltech.edu/datasets/cub_200_2011/}.

\bibitem[Wald et~al.(2021)Wald, Feder, Greenfeld, and Shalit]{dg_calib}
Wald, Y., Feder, A., Greenfeld, D., and Shalit, U.
\newblock On calibration and out-of-domain generalization.
\newblock \emph{NeurIPS}, 2021.
\newblock URL \url{https://arxiv.org/abs/2102.10395}.

\bibitem[{Wang} et~al.(2021){Wang}, {Lan}, {Liu}, {Ouyang}, {Qin}, {Lu},
  {Chen}, {Zeng}, and {Yu}]{dg_survey_2}
{Wang}, J., {Lan}, C., {Liu}, C., {Ouyang}, Y., {Qin}, T., {Lu}, W., {Chen},
  Y., {Zeng}, W., and {Yu}, P.~S.
\newblock Generalizing to unseen domains: A survey on domain generalization.
\newblock \emph{arXiv}, 2021.
\newblock URL \url{https://arxiv.org/abs/2103.03097}.

\bibitem[Williams et~al.(2017)Williams, Nangia, and Bowman]{williams2017broad}
Williams, A., Nangia, N., and Bowman, S.~R.
\newblock A broad-coverage challenge corpus for sentence understanding through
  inference.
\newblock \emph{ACL}, 2017.
\newblock URL \url{https://aclanthology.org/N18-1101/}.

\bibitem[Woodward(2005)]{woodward_making}
Woodward, J.
\newblock \emph{Making things happen: A theory of causal explanation}.
\newblock Oxford university press, 2005.

\bibitem[Wortsman et~al.(2022)Wortsman, Ilharco, Gadre, Roelofs, Gontijo-Lopes,
  Morcos, Namkoong, Farhadi, Carmon, Kornblith, et~al.]{wortsman2022model}
Wortsman, M., Ilharco, G., Gadre, S.~Y., Roelofs, R., Gontijo-Lopes, R.,
  Morcos, A.~S., Namkoong, H., Farhadi, A., Carmon, Y., Kornblith, S., et~al.
\newblock Model soups: averaging weights of multiple fine-tuned models improves
  accuracy without increasing inference time.
\newblock In \emph{ICML}, 2022.
\newblock URL \url{https://arxiv.org/abs/2203.05482}.

\bibitem[Xiao et~al.(2020)Xiao, Engstrom, Ilyas, and Madry]{xiao2020noise}
Xiao, K., Engstrom, L., Ilyas, A., and Madry, A.
\newblock Noise or signal: The role of image backgrounds in object recognition.
\newblock \emph{arXiv}, 2020.
\newblock URL \url{https://arxiv.org/abs/2006.09994}.

\bibitem[Yang et~al.(2023)Yang, Zhang, Katabi, and Ghassemi]{yang2023change}
Yang, Y., Zhang, H., Katabi, D., and Ghassemi, M.
\newblock Change is hard: A closer look at subpopulation shift.
\newblock \emph{arXiv}, 2023.
\newblock URL \url{https://arxiv.org/pdf/2302.12254.pdf}.

\bibitem[Yao et~al.(2022)Yao, Wang, Li, Zhang, Liang, Zou, and Finn]{lisa}
Yao, H., Wang, Y., Li, S., Zhang, L., Liang, W., Zou, J., and Finn, C.
\newblock Improving out-of-distribution robustness via selective augmentation.
\newblock \emph{arXiv}, 2022.
\newblock URL \url{https://arxiv.org/abs/2201.00299}.

\bibitem[Zhang et~al.(2018)Zhang, Cisse, Dauphin, and Lopez-Paz]{mixup}
Zhang, H., Cisse, M., Dauphin, Y.~N., and Lopez-Paz, D.
\newblock mixup: Beyond empirical risk minimization.
\newblock In \emph{ICLR}, 2018.
\newblock URL \url{https://arxiv.org/abs/1710.09412}.

\bibitem[Zhang et~al.(2022{\natexlab{a}})Zhang, Lopez-Paz, and Bottou]{leon}
Zhang, J., Lopez-Paz, D., and Bottou, L.
\newblock Rich feature construction for the optimization-generalization
  dilemma.
\newblock In \emph{International Conference on Machine Learning}, pp.\
  26397--26411. PMLR, 2022{\natexlab{a}}.

\bibitem[Zhang et~al.(2023)Zhang, Li, and Wang]{zhang2023semi}
Zhang, L., Li, J.-F., and Wang, W.
\newblock Semi-supervised domain generalization with known and unknown classes.
\newblock In \emph{Thirty-seventh Conference on Neural Information Processing
  Systems}, 2023.

\bibitem[Zhang et~al.(2021)Zhang, Marklund, Dhawan, Gupta, Levine, and
  Finn]{zhang2021adaptive}
Zhang, M., Marklund, H., Dhawan, N., Gupta, A., Levine, S., and Finn, C.
\newblock Adaptive risk minimization: Learning to adapt to domain shift.
\newblock \emph{NeurIPS}, 2021.
\newblock URL \url{https://arxiv.org/abs/2007.02931}.

\bibitem[Zhang et~al.(2022{\natexlab{b}})Zhang, Sohoni, Zhang, Finn, and
  R{\'e}]{cnc}
Zhang, M., Sohoni, N.~S., Zhang, H.~R., Finn, C., and R{\'e}, C.
\newblock Correct-n-contrast: A contrastive approach for improving robustness
  to spurious correlations.
\newblock \emph{arXiv preprint arXiv:2203.01517}, 2022{\natexlab{b}}.

\bibitem[{Zheran Liu} et~al.(2021){Zheran Liu}, {Haghgoo}, {Chen},
  {Raghunathan}, {Koh}, {Sagawa}, {Liang}, and {Finn}]{jtt}
{Zheran Liu}, E., {Haghgoo}, B., {Chen}, A.~S., {Raghunathan}, A., {Koh},
  P.~W., {Sagawa}, S., {Liang}, P., and {Finn}, C.
\newblock Just train twice: Improving group robustness without training group
  information.
\newblock \emph{arXiv}, 2021.
\newblock URL \url{https://arxiv.org/abs/2107.09044}.

\bibitem[Zhou et~al.(2022{\natexlab{a}})Zhou, Liu, Qiao, Xiang, and
  Loy]{dg_survey}
Zhou, K., Liu, Z., Qiao, Y., Xiang, T., and Loy, C.~C.
\newblock Domain generalization: A survey.
\newblock \emph{IEEE PAMI}, 2022{\natexlab{a}}.
\newblock URL \url{https://arxiv.org/abs/2103.02503}.

\bibitem[Zhou et~al.(2022{\natexlab{b}})Zhou, Lin, Zhang, and
  Zhang]{pmlr-v162-zhou22e}
Zhou, X., Lin, Y., Zhang, W., and Zhang, T.
\newblock Sparse invariant risk minimization.
\newblock In \emph{Proceedings of the 39th International Conference on Machine
  Learning}, 2022{\natexlab{b}}.
\newblock URL \url{https://proceedings.mlr.press/v162/zhou22e.html}.

\end{thebibliography}
\bibliographystyle{icml2024}

\newpage
\appendix
\onecolumn

\section{Further Related Work}
\label{app:related_work}

The literature on OOD and domain generalization spans a decade and comprises a vast amount of works.
In the review below, we survey (i) some of the major milestones of OOD generalization research, (ii) advances in the sub-problem of sub-population shift, (iii) the multifarious connections between OOD generalization and causal inference, (iv) efforts to learn domains, sub-populations, or environments from pooled collections of training examples previous to our XRM proposal, and (v) their limitations in terms of annotation requirements and impossibility results.

\begin{enumerate}[label=(\roman*),leftmargin=2em]
    \item The first works in \emph{OOD and domain generalization} proposed algorithms that summarize each domain as a kernel mean embedding of the respective distribution of inputs~\citep{blanchard, pmlr-v28-muandet13}; these were later extended to the realm of deep neural networks~\citep{zhang2021adaptive}.
    One common avenue towards domain generalization is to learn a predictor where the feature representation has the same distribution across domains~\citep{sun2016deep, dann}.
    Another major strategy is to enforce learning a richer feature space~\citep{leon}, which can be done by combining the weights of multiple models with different hyper-parameter configurations~\citep{cha2021swad, rame2022diverse, wortsman2022model, lin2023spurious}, or biasing training to make them disagree with each other~\citep{lff, pagliardini2022agree, lee2022diversify}.
    Learning from combinations of examples, by means of mixup~\citep{mixup}, is also a promising route to diminish the impact of spurious correlations~\citep{lisa, giannone2022just}.
    All in all, there are multiple frameworks that evaluate dozens of OOD generalization algorithms across a variety of benchmark datasets, such as DomainBed~\citep{domainbed} and WILDS~\citep{wilds}.
    We recommend the reader to consult recent surveys~\citep{dg_survey, dg_survey_2} for a taxonomy of the vast array algorithms on offer.
    \item \emph{Sub-population shift} is a particular type of OOD generalization problem, where environments are direct annotations of a spurious attribute, and one can assume that the test domain will be equal a subdistribution---group---of the training data.
    The gold-standard for addressing sub-population shifts is group distributionally robust optimization~\citep[GroupDRO]{dro}.
    Group subsampling and reweighting schemes, albeit simple, also provide state-of-the-art accuracy~\citep{subg}.
    To achieve good performance, it is known that it suffices to finetune the last layer of a deep neural network with a small training set with balanced groups~\citep{izmailov2022feature}.
    The framework of SubpopBench compares twenty algorithms for sub-population shift across a dozen benchmark datasets~\citep{yang2023change}.
    \item The goal of OOD/domain generalization can be understood as finding predictors invariant across a family of relevant environments~\citep{irm}.
    This establishes an intimate link between OOD generalization and causality under the interventionist account, where causation is defined as invariance across interventions~\citep{woodward_making}.
    A pioneering method attacks the problem OOD generalization as finding invariant causal predictors~\citep[ICP]{peters2016causal}.
    The framework of invariant risk minimization (IRM)~\cite{irm} extends ICP to deep neural networks, advocating the invariance principle of ``finding a feature representation such that the optimal classifier matches across environments''.
    Researchers have proposed multiple variants of the original IRM formulation, with notable examples being risk extrapolation~\citep[vREX]{krueger2021out} and sparse risk minimization~\citep{pmlr-v162-zhou22e}.
    The IRM framework has found multiple applications, with fair face recognition~\cite{ma2023invariant} being a recent example.
    \item The main factor limiting the application of OOD generalization and sub-population shift machinery is their requirement of domain, environment, or group annotations.
    Unfortunately, these are resource-intensive to obtain and are limited by human annotators' biases, as the biases they identify may not align with those learned by models, and vice versa~\citep{bell2023simplicity}.

    Consequently, a wide array of methods has been recently proposed to estimate these annotations from pooled collections of training data~\citep{labonte2023towards,zhang2023semi,tsirigotis2023group}.
    Among them, learning from failure~\citep[LfF]{lff} learns a biased network, and a final network that focuses on the examples misclassified by the biased network.
    Environment inference for invariant learning~\citep[EIIL]{eiil} searches for an environmental partition that violates the IRM principle. 
    Just-train-twice~\citep[JTT]{jtt} trains one first network for a few iterations, and a final network to focus more on the examples from the first network.
    Correct and Contrast~\citep[CNC]{cnc} leverages ERM failures and contrastive learning to learn a robust representation.
    Automatic feature reweighting~\citep[AFR]{qiu2023simple} learns a first network for a few iterations, and then fine-tunes the last layer to focus on mistakes.
    Learning to split~\citep[LS]{ls} and adversarial re-weighted learning~\citep[ARL]{arl} implement adversarial games to find a split of the training data inducing maximum out-of-sample error. 
    Bias amplification~\citep[BAM]{bam} incorporates per-example ``slack variables'' to absorb the fast learning of spurious correlations. 
    No subclass left behind~\citep[GEORGE]{sohoni2020no} clusters the hidden representation of a neural network to construct different environments.
    \cite{teney2021unshuffling} manually identify variables to stratify pooled collections of training examples into environments.
    In the context of label noise robustness, the prior CrossSplit~\citep{kim2023crosssplit} uses a similar approach to that of XRM. CrossSplit uses a pair of networks trained on two disjoint parts of the training data and uses label correction to prevent memorization of noisy (mis-labelled) examples. While confidently mistaken cross-predictions are indicators of a model’s memorization of a noisy label in the context of the CrossSplit paper, they indicate a model’s reliance on spurious correlations in our work.
    In the context of graph OOD generalization, GALA \cite{chen2024does} was proposed as a framework for graph OOD generalization that uses an assistant ERM model to identify invariant subgraphs through agreement and disagreement with proxy predictions.

    \item One important note to the environment discovery methods described above is that they still require group annotations in a validation set, used for selecting a model with good worst-group-accuracy.
    In the complete absence of environment annotations, learning invariant predictors is an impossible task in its full generality~\citep{chen2022pareto, lin2022zin, pmlr-v202-tan23b, chen2023rethinking}.
    Because we have proposed XRM as an alternative to surmount such daunting task, the next section provides intuitions to identify success and failure cases of our method.
\end{enumerate}

\section{Experimental Details}
\label{app:exp}

\subsection{Sub-population Shift Experiments}

For the results in~\cref{tab:xrm_effectiveness}, we follow standard experimental protocol as in~\citep{yang2023change}.
For model selection, we adhere to the standard practice of using the worst-group-accuracy.
We try 10 different hyper-parameter combinations detailed in~\cref{app:hp} with one random seed.
We select the hyper-parameter combination and early-stopping iteration yielding maximal validation worst-group-accuracy (or, in the absence of groups, worst-class-accuracy).
Next, we repeat the experiment 10 times with different seeds.
Regardless of how training and validation groups are discovered, we always report test worst-group-accuracy over the \textbf{human} group annotations provided by each dataset.

We consider six standard datasets.
These are the four image datasets Waterbirds~\citep{wah2011caltech}, CelebA~\citep{celeba}, MetaShift~\citep{liang2022metashift}, and ImageNetBG~\citep{xiao2020noise}; and the two natural language datasets MultiNLI~\citep{williams2017broad} and CivilComments~\citep{borkan2019nuanced}.
For CelebA, predictors map pixel intensities into a binary ``blonde/not-blonde'' label. No individual face characteristics, landmarks, keypoints, facial mapping, metadata, or any other information was used to train our CelebA predictors.
Image datasets use a pretrained ResNet-50~\citep{he2016deep}, frozen for \method{} experiments. Text datasets use a pretrained BERT~\citep{devlin2018bert}. The linear layers on the top of the pretrained models are initialized at zero.
All images are resized and center-cropped to $224 \times 224$ pixels, and undergo no data augmentation.
We use SGD with momentum $0.9$ to learn from image datasets unless otherwise mentioned, and we employ AdamW~\citep{loshchilov2017decoupled} with default $\beta_1=0.9$ and $\beta_2=0.999$ for text benchmarks.
As for the OOD generalization algorithms, we consider ERM, group distributionally robust optimization \citep[GroupDRO]{dro}, group re-weighting~\citep[RWG]{japkowicz2000class}, and group sub-sampling~\citep[SUBG]{subg}.

We also conduct experiments on ColorMNIST~\citep{irm}, but keep a strict protocol.
More specifically, we set \emph{both} training and validation data to contain two environments, with $0.8$ and $0.9$ label-color correlation, while the test environment shows $0.1$ label-color correlation.
This contrasts \citet{irm}, who used the test environment for model selection.
We train a three-layer fully-connected network with layer sizes $[2 * 14 * 14, 300, 300, 2]$ and use ReLU as the activation function. The network is optimized using the Adam optimizer with a learning rate of $1e-3$, and default parameters $\beta_1=0.9$ and $\beta_2=0.999$.
We train all algorithms for a number of iterations that allows convergence within a reasonable compute budget.
Full results with error bars are reported in~\cref{tab:xrm_full}.
\begin{table}[h]
\caption{Worst-group-accuracies and standard deviations for all datasets, algorithms, and annotations over 10 trials.}
\vspace{-0.5cm}
\label{tab:xrm_full}
\begin{center}
\resizebox{\textwidth}{!}{%
\begin{NiceTabular}{lcccccccccccc}
\CodeBefore
\rectanglecolor{papercolor!10}{2-4}{11-4}
\rectanglecolor{papercolor!10}{2-7}{11-7}
\rectanglecolor{papercolor!10}{2-10}{11-10}
\rectanglecolor{papercolor!10}{2-13}{11-13}
\Body
\toprule
&
\multicolumn{3}{c}{\textbf{ERM}} &
\multicolumn{3}{c}{\textbf{GroupDRO}} &
\multicolumn{3}{c}{\textbf{RWG}} &
\multicolumn{3}{c}{\textbf{SUBG}} \\
\midrule
&	None	&	Human	&	\textbf{XRM}	&	None	&	Human	&	\textbf{XRM}	&	None	&	Human	&	\textbf{XRM}	&	None	&	Human	&	\textbf{XRM}	\\
\midrule																						
\textbf{Waterbirds}     & 66.4 \scriptsize$\pm$ 0.9 & 66.4 \scriptsize$\pm$ 0.9 & 66.4 \scriptsize$\pm$ 0.9 & 67.3 \scriptsize$\pm$ 0.6 & 86.5 \scriptsize$\pm$ 0.5 & 88.1 \scriptsize$\pm$ 0.9 & 66.8 \scriptsize$\pm$ 0.5 & 87.3 \scriptsize$\pm$ 0.3 & 82.0 \scriptsize$\pm$ 1.0 & 62.1 \scriptsize$\pm$ 1.5 & 87.1 \scriptsize$\pm$ 1.1 & 77.7 \scriptsize$\pm$ 2.7 \\
\textbf{CelebA}         & 54.3 \scriptsize$\pm$ 2.7 & 55.1 \scriptsize$\pm$ 1.8 & 58.6 \scriptsize$\pm$ 3.8 & 68.4 \scriptsize$\pm$ 2.1 & 88.3 \scriptsize$\pm$ 2.1 & 89.1 \scriptsize$\pm$ 1.3 & 64.7 \scriptsize$\pm$ 1.1 & 82.6 \scriptsize$\pm$ 1.8 & 81.1 \scriptsize$\pm$ 2.1 & 65.9 \scriptsize$\pm$ 2.7 & 83.9 \scriptsize$\pm$ 2.8 & 81.1 \scriptsize$\pm$ 2.3 \\
\textbf{MultiNLI}       & 67.9 \scriptsize$\pm$ 2.0 & 72.0 \scriptsize$\pm$ 1.2 & 69.1 \scriptsize$\pm$ 1.9 & 68.6 \scriptsize$\pm$ 1.4 & 73.4 \scriptsize$\pm$ 4.8 & 72.1 \scriptsize$\pm$ 1.0 & 66.9 \scriptsize$\pm$ 1.8 & 71.3 \scriptsize$\pm$ 1.7 & 70.2 \scriptsize$\pm$ 1.3 & 69.7 \scriptsize$\pm$ 1.4 & 52.4 \scriptsize$\pm$ 25.8 & 72.0 \scriptsize$\pm$ 1.2 \\
\textbf{CivilComments}  & 67.2 \scriptsize$\pm$ 1.3 & 74.0 \scriptsize$\pm$ 0.5 & 64.7 \scriptsize$\pm$ 9.8 & 66.6 \scriptsize$\pm$ 1.1 & 73.8 \scriptsize$\pm$ 0.6 & 72.2 \scriptsize$\pm$ 0.8 & 67.2 \scriptsize$\pm$ 1.3 & 73.4 \scriptsize$\pm$ 0.5 & 72.2 \scriptsize$\pm$ 0.4 & 65.4 \scriptsize$\pm$ 1.9 & 71.1 \scriptsize$\pm$ 1.4 & 44.6 \scriptsize$\pm$ 4.2 \\
\textbf{ColorMNIST}     & 10.0 \scriptsize$\pm$ 0.0 & 10.1 \scriptsize$\pm$ 0.3 & 11.3 \scriptsize$\pm$ 0.9 & 9.9  \scriptsize$\pm$ 0.0 & 10.1 \scriptsize$\pm$ 0.2 & 69.7 \scriptsize$\pm$ 1.1 & 10.0 \scriptsize$\pm$ 0.0 & 10.3 \scriptsize$\pm$ 0.4 & 69.4 \scriptsize$\pm$ 1.8 & 10.0 \scriptsize$\pm$ 0.0 & 9.9  \scriptsize$\pm$ 0.3 & 65.2 \scriptsize$\pm$ 2.2 \\
\textbf{MetaShift}      & 64.2 \scriptsize$\pm$ 1.0 & 69.8 \scriptsize$\pm$ 3.3 & 71.7 \scriptsize$\pm$ 2.1 & 72.5 \scriptsize$\pm$ 3.0 & 80.3 \scriptsize$\pm$ 0.9 & 77.5 \scriptsize$\pm$ 1.0 & 72.3 \scriptsize$\pm$ 2.5 & 76.3 \scriptsize$\pm$ 2.1 & 74.2 \scriptsize$\pm$ 2.0 & 69.7 \scriptsize$\pm$ 3.7 & 77.0 \scriptsize$\pm$ 3.4 & 77.6 \scriptsize$\pm$ 1.4 \\
\textbf{ImagenetBG}     & 78.4 \scriptsize$\pm$ 1.2 & 78.4 \scriptsize$\pm$ 1.3 & 79.0 \scriptsize$\pm$ 0.8 & 78.3 \scriptsize$\pm$ 0.9 & 78.2 \scriptsize$\pm$ 0.9 & 78.2 \scriptsize$\pm$ 1.2 & 78.3 \scriptsize$\pm$ 1.5 & 79.0 \scriptsize$\pm$ 1.5 & 78.8 \scriptsize$\pm$ 1.4 & 79.5 \scriptsize$\pm$ 1.1 & 78.8 \scriptsize$\pm$ 1.2 & 77.2 \scriptsize$\pm$ 1.2 \\
\bottomrule
\end{NiceTabular}%
}
\end{center}
\end{table}

For the experiment on CIFAR-10 \citep{cifar}, we train a VGG-16 model \citep{simonyan2014very} using SGD with a learning rate of $1e-2$ and a momentum of $0.9$.

\subsection{\textsc{DomainBed} Experiments}
\label{app:domainbed}
For the results in~\cref{tab:domainbed_summary} and~\cref{tab:domainbed_full}, we adhere to the original codebase from \textsc{DomainBed}~\citep{domainbed}.
Each dataset is used to train a model on all but one environment, which is held out as the test domain.
This process is repeated for each possible environment as the test domain, resulting in multiple training and testing splits for each dataset.
We experiment with ERM and the CORAL algorithm as it is the best performing single-model (non-ensembling) method according to the \textsc{DomainBed} suite.
For each triplet of (dataset, method, test environment), we sweep over 10 different hyper-parameter combinations detailed in~\cref{app:hp}.
We perform model selection based on the \textit{average accuracy over the validation environments}, which is referred to as the \textit{`training domain validation set'} in the \textsc{DomainBed} paper.
In~\cref{tab:domainbed_full}, we report the results for each environment when selected as the test environment. Additionally, we report the average and worst environment test accuracies.
In settings without environment annotations (i.e., ERM), we combine all training environments and then split into one training and one validation set.
In those cases with annotations, whether human-annotated or discovered by XRM, each training environment is divided into as many training and validation sets as the number of environments.
\begin{table}[h]
\caption{Full results for the \textsc{DomainBed} suite.}
\label{tab:domainbed_full}
\begin{subtable}{\textwidth}
\centering
\resizebox{0.6\textwidth}{!}{%
\begin{NiceTabular}{lcccc|cc}
\CodeBefore
\rectanglecolor{papercolor!10}{2-7}{5-7}
\Body
\toprule                        
\textbf{VLCS} & C   & L & S & V & Avg   & Worst \\
\midrule                        
\textbf {ERM} (None)    & 96.70 & 64.85 & 74.20 & 76.15 & 77.97 & 64.85 \\
\textbf{CORAL} (Human)  & 97.35 & 65.00 & 72.80 & 76.35 & 77.87 & 65.00 \\
\textbf{CORAL} (XRM)    & 95.55 & 66.15 & 72.45 & 76.50 & 77.66 & 66.15 \\
\bottomrule
\end{NiceTabular}}%
\end{subtable}
\begin{subtable}{\textwidth}
\centering
\vspace{0.35cm}
\resizebox{0.6\textwidth}{!}{%
\begin{NiceTabular}{lcccc|cc}
\CodeBefore
\rectanglecolor{papercolor!10}{2-7}{5-7}
\Body
\toprule                        
\textbf{PACS}   & A & C & P & S & Avg   & Worst \\
\midrule                        
\textbf {ERM} (None)    & 84.65 & 80.65 & 95.55 & 72.55 & 83.35 & 72.55 \\
\textbf{CORAL} (Human)  & 84.90 & 80.75 & 96.60 & 77.70 & 84.98 & 77.70 \\
\textbf{CORAL} (XRM)    & 81.90 & 77.30 & 96.90 & 79.15 & 83.81 & 77.30 \\
\bottomrule
\end{NiceTabular}}%
\end{subtable}
\begin{subtable}{\textwidth}
\centering
\vspace{0.35cm}
\resizebox{0.6\textwidth}{!}{%
\begin{NiceTabular}{lcccc|cc}
\CodeBefore
\rectanglecolor{papercolor!10}{2-7}{5-7}
\Body
\toprule                        
\textbf{OfficeHome} & A & C & P & R & Avg   & Worst \\
\midrule                        
\textbf {ERM} (None)    & 59.50 & 52.25 & 74.15 & 76.00 & 65.47 & 52.25 \\
\textbf{CORAL} (Human)  & 64.00 & 53.55 & 76.15 & 77.25 & 67.73 & 53.55 \\
\textbf{CORAL} (XRM)    & 61.80 & 53.90 & 74.85 & 77.50 & 67.01 & 53.90 \\
\bottomrule
\end{NiceTabular}}%
\end{subtable}
\begin{subtable}{\textwidth}
\centering
\vspace{0.35cm}
\resizebox{0.6\textwidth}{!}{%
\begin{NiceTabular}{lcccc|cc}
\CodeBefore
\rectanglecolor{papercolor!10}{2-7}{5-7}
\Body
\toprule                        
\textbf{TerraIncognita} & L100  & L38   & L43   & L46   & Avg   & Worst \\
\midrule                        
\textbf {ERM} (None)    & 54.80 & 42.30 & 56.40 & 34.60 & 47.02 & 34.60 \\
\textbf{CORAL} (Human)  & 58.20 & 39.25 & 59.45 & 37.15 & 48.51 & 37.15 \\
\textbf{CORAL} (XRM)    & 59.20 & 45.10 & 56.10 & 38.00 & 49.60 & 38.00 \\
\bottomrule
\end{NiceTabular}}%
\end{subtable}
\begin{subtable}{\textwidth}
\centering
\vspace{0.35cm}
\resizebox{0.6\textwidth}{!}{%
\begin{NiceTabular}{lcccccc|cc}
\CodeBefore
\rectanglecolor{papercolor!10}{2-9}{5-9}
\Body
\toprule                                
\textbf{DomainNet}  & clip  & info  & paint & quick & real  & sketch    & Avg   & Worst \\
\midrule                                
\textbf {ERM} (None)    & 47.40 & 14.75 & 37.45 & 9.30  & 42.10 & 39.15 & 31.69 & 9.30 \\
\textbf{CORAL} (Human)  & 60.05 & 20.25 & 47.90 & 13.25 & 59.95 & 50.45 & 41.97 & 13.25 \\
\textbf{CORAL} (XRM)    & 50.40 & 16.80 & 42.30 & 11.60 & 50.40 & 43.70 & 35.87 & 11.60 \\
\bottomrule
\end{NiceTabular}}%
\end{subtable}
\end{table}

\subsection{Dominoes Experiments}

We conduct an additional experiment on the dataset of Dominoes \citep{geirhos2020shortcut}. Dominoes is an image dataset in which images from two other datasets are concatenated. Particularly, the top half of an image shows MNIST digits from classes \{0, 1\}. The bottom half shows images from classes \{coat, dress\} of FashionMNIST in the MF (MNIST-MNISTFashion) version. And in the MC (MNIST-CIFAR) version, the bottom half shows images from classes \{car, truck\} of CIFAR-10. We followed a setup as described in \cite{kirichenko2022last} (except that we used a smaller ConvNet) where the spurious MNIST part is strongly correlated (99\% of the time) with the labels in the training data but drops to 50\% (random) in the validation and test sets. Meanwhile, the core feature (either FashionMNIST or CIFAR) is always fully correlated with the labels. We compare three methods on these datasets: the Vanilla ERM, GroupDRO) using the ground-truth annotations\footnote{annotations indicate whether the MNIST half matches the label or not}, and GroupDRO with XRM-inferred annotations.
The results, shown in Table \ref{tab:dominoes}, suggest that the group annotations inferred by XRM are as effective as the ground-truth annotations.

\begin{figure}[t!]
    \centering
    \includegraphics[width=0.90\textwidth]{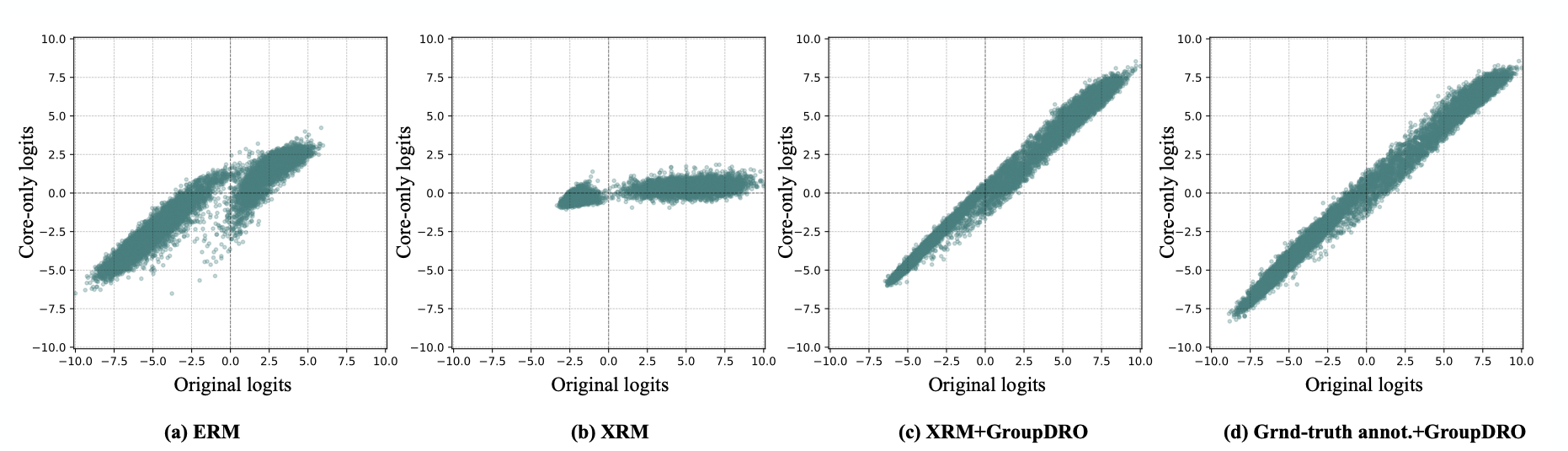}
    \caption{
    \textbf{Logit scatter plots for models trained on Dominoes-MF.} The x-axis shows the logits with the original training examples, while the y-axis displays the logits for the same training examples with the MNIST part removed. These plots help reveal the extent of reliance on the spurious feature (MNIST digit) versus the core feature (FashionMNIST or Cifar). \textbf{a)} Vanilla ERM Model: Strong reliance on the spurious MNIST digit feature and partial dependence on the core feature. \textbf{b)} XRM Model: Strong reliance on the spurious MNIST digit. \textbf{c)} GroupDRO Model with XRM-Inferred annotations: point mostly on the diagonal suggesting invariance to the spurious feature, focusing only on the core feature. \textbf{d)} GroupDRO with ground-truth annotations for reference. We highlight that for XRM, it is desirable to only rely on the spurious feature since this will then enable the subsequent GroupDRO to learn the invariant core feature.}
    \vspace{-0.4cm}
    \label{fig:dominoes}

\end{figure}

\begin{table}[ht]
\centering
\caption{Worst-group accuracy of three methods on two Dominoes datasets. Dominoes MF and MC are concatenation of MNIST digits with FashionMNIST and Cifar-10, respectively. The MNIST digit is spurious with 99\% correlation in the training set and random correlation in the validation and test sets. The core feature (either FashionMNIST or Cifar) is however always fully correlated with the labels. GT denotes ground-truth group annotations.}
\label{tab:dominoes}
\begin{tabular}{lccc}
\toprule
 & ERM& XRM+GroupDRO& GT+GroupDRO \\
\midrule
Dominoes MF & 50.74 & 86.68 & 85.28 \\
Dominoes MC & 48.30 & 68.78 & 69.43\\

\bottomrule
\end{tabular}
\end{table}

To further compare these methods, in Figure \ref{fig:dominoes}, we provide a scatter plot of the model logits for each method. Specifically, we feed two versions of all training examples to a fully trained model and store two sets of logits. One set is the result of the original training examples, and the other set is obtained with the MNIST part removed (replaced by average).
\vspace{-0.2cm}
\begin{itemize}[itemsep=0pt]
  \item For an ERM model, the logits vary mostly along the x-axis and less along the y-axis, suggesting that the model is mostly relying on the spurious feature but also partially on the core feature.
  \item An XRM model varies mostly along the x-axis and is almost invariant along the y-axis, indicating that the model strongly relies on the spurious feature alone and hence can effectively split the data into two environments according to the spurious correlation.
  \item For a model trained with GroupDRO on XRM-annotations, points are almost all on the diagonal, suggesting that removing the spurious feature does not change the logits much, implying that the model is invariant to the spurious feature and only relies on the core feature.
  \item A model trained with GroupDRO using the ground-truth annotations shows similar results, proving it mainly relies on the core feature.
\end{itemize}

\subsection{Hyperparameter Sensitivity Analysis of \method{} on Waterbirds}

\begin{figure}[t!]
    \centering
    \includegraphics[width=0.90\textwidth]{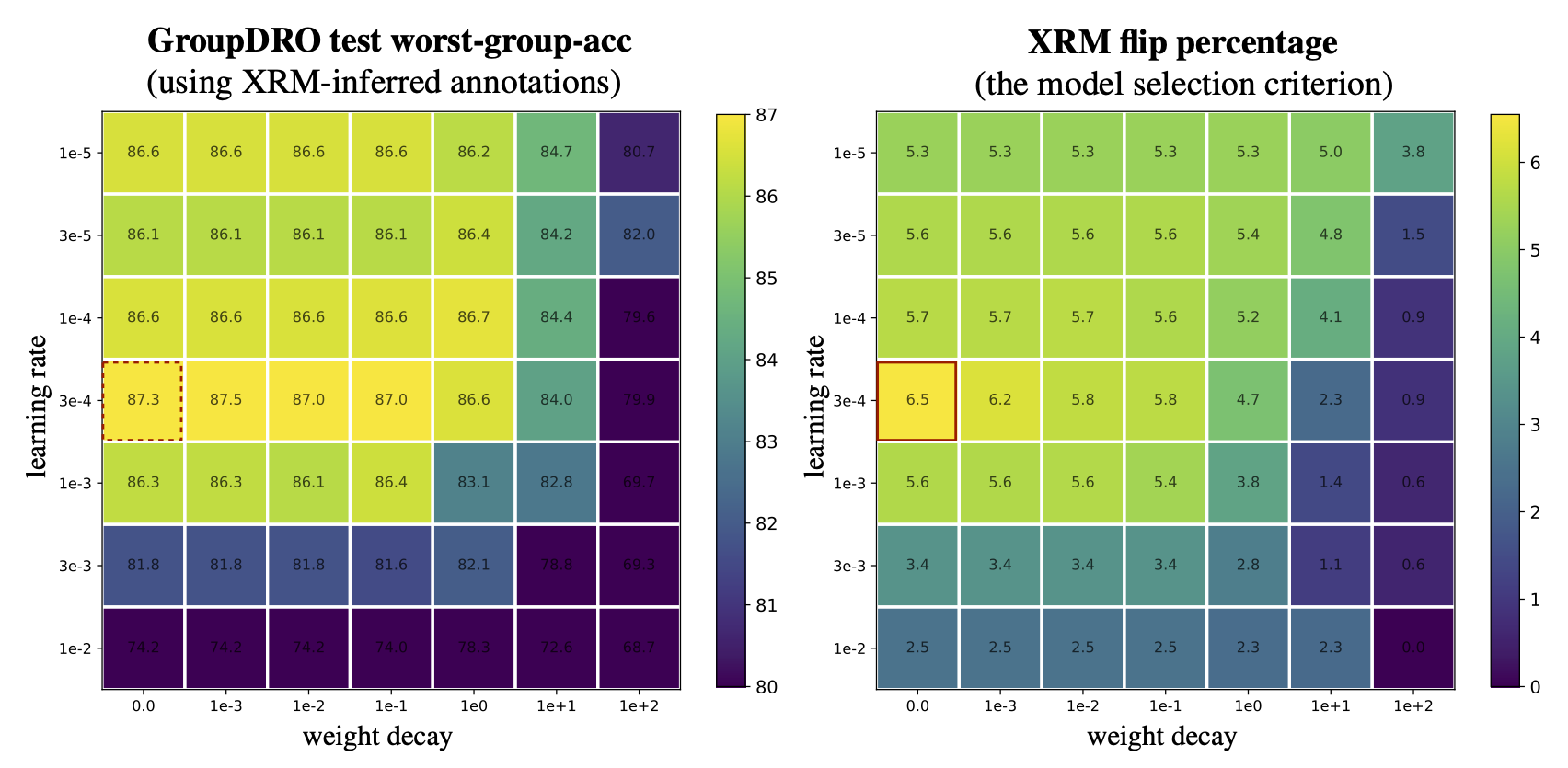}
    \caption{
    \textbf{Hyperparameter sensitivity analysis of XRM on the Waterbirds.} Here we evaluate the sensitivity of XRM's performance to variations in learning rate and weight decay hyperparameters, while keeping the batch size fixed at 512. \textbf{Left:} worst-group-acc of a GroupDRO model trained with XRM annotations. Each cell within the grid represents a hyperparameter combination for XRM, with the color intensity indicating the test worst-group-accuracy. \textbf{Right:} the percentage of labels flipped by XRM for the corresponding hyperparameter combination, serving as the model selection criterion. \textbf{Analysis:} The best XRM model is chosen according to the highest flip percentage (red square), which resulted in an accuracy of 87.3\% (dashed red square). However, the plot on the left shows that even neighboring cells with very different hyperparameters can still lead to near-optimal performance.
    }
    \label{fig:hp_sensitivity}
\end{figure}

Here, we conduct a hyperparameter sensitivity experiment on the Waterbirds dataset and report the results in Figure \ref{fig:hp_sensitivity}. We fix the batch-size at 512 and vary the learning rate and weight decay coefficient to see how \method{} performs with different hyperparameters. This grid of outcomes across different learning rate and weight decay settings shows that \method{} works well across a wide range of hyperparameter combinations.

For each combination of learning rate and weight decay (each cell in the grid), we fully train an \method{} model, get the group annotations, and then use GroupDRO with these inferred annotations. The hyperparameters for GroupDRO kept the same throughout all experiments. Hence, any variability in performance is only attributed to the \method{} phase.

\subsection{Can XRM Handle Settings with More Than Two True Underlying Environments?}
\begin{figure}[t!]
    \centering
    \includegraphics[width=0.75\textwidth]{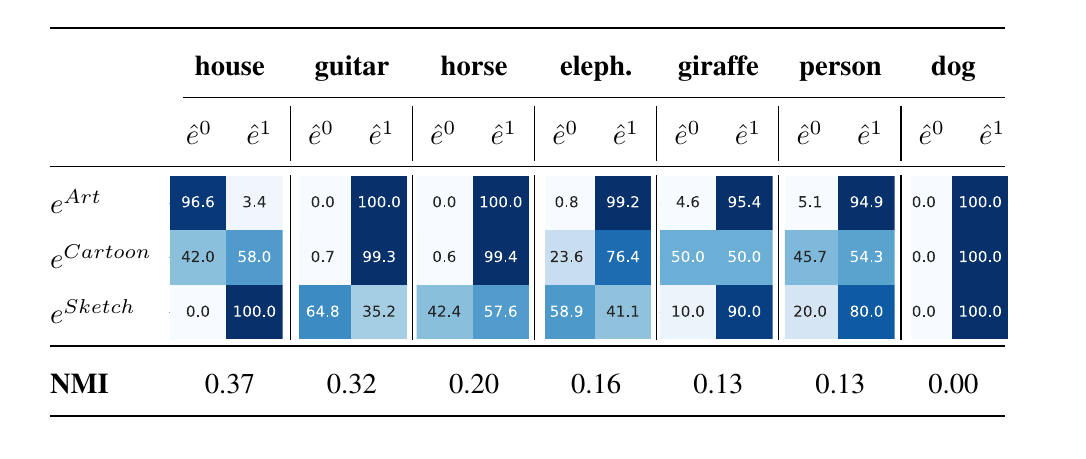}
\caption{\textbf{Confusion matrices of XRM-inferred environments vs. human-annotated environments for the PACS dataset.}
Each row corresponds to one of the human-annotated environments, $e \in \{$Art, Cartoon, Sketch$\}$. Columns represent the two environments, $\hat{e}^{0 / 1}$ inferred by XRM for each class. Entries in the matrices indicate the percentage of examples from each human-annotated environment that XRM grouped into each of its two inferred environments. Normalized Mutual Information (NMI), a metric between zero and one, quantifies the alignment between the inferred and annotated environments. \textbf{Analysis:} For certain classes (e.g., ``house''), the XRM-inferred environments align well with human annotations, while for other classes (e.g., ``dog''), no alignment is observed. This is acceptable as long as the inferred environments can be used for invariant learning, which appears to be the case. XRM's goal is to find environments suitable for subsequent invariant learning, rather than necessarily recovering the original environments. As a reference point, a random splitter would result in an NMI of zero on all classes, and a subsequent invariant learning method based on random split cannot perform better than ERM. Worst-group-accuracy of ERM is 72.55 while CORAL with Human: 77.70, \ XRM: 77.30, \ random envs: 71.60.}
\label{fig:confusion}
\end{figure}

XRM is always presented with pooled data with no explicit annotations. Regardless of how many human-annotation environments are in the pooled data, XRM always splits the data into 2 environments (2 groups per label). For example, all the datasets in table \ref{fig:confusion}, have more than two human-annotated environments. That is while XRM split the examples into two environments.

Whether two environments suffice—when constructed appropriately—is in fact a fascinating question that we have pondered about for a while. Our experiments, especially those on the DomainBed, show that although XRM splits the data into a smaller number of groups than the human annotations, it is successful in forming sufficiently varied groups to match the performance with human annotations.

To illustrate, take one of the experiments with the PACS dataset where the “Photo” environment is left for testing and the other three environments of “Art”, “Cartoon”, and “Sketch” are pooled together and presented to XRM. Now XRM’s objective is to split the dataset into environments that lead to learning an invariant predictor, not to recover the original environment. XRM’s splitting is done according to the ‘hardness’ of examples and might or might not align with the human annotations. To see that, we visualize a per-class confusion matrix for the environment discovery of XRM. We also quantify the alignment between XRM environments and human-annotated ones using Normalized-Mutual-Information (NMI), a higher NMI means more alignment.

We observe that for some of the classes (e.g., the “House” class) there is a rather strong alignment. However, for some other classes (e.g., the “Dog” class), there is no alignment between human-annotated environments and those of XRM. That is absolutely fine as long as the inferred environments can be used for invariant learning and that appears to be the case.

As a reference point, a random splitter would lead to an NMI of zero on all classes and a subsequent invariant 
learning method cannot do better than an ERM with no environment annotations. 

\subsection{\method{} Model Selection}

To determine the best hyper-parameter combination for each experiment, we run \method{} with 10 different hyper-parameter combinations detailed in~\cref{app:hp} and one random seed.
We then compute the percentage of flipped labels that appear at the last iteration for each combination.
The combination with the highest percentage is selected as the best. We disregard rare cases in which a hyper-parameter combination yields degenerate grouping where every example in one class in misclassified.
Next, we repeat the experiment 10 times with different seeds to obtain 10 sets of inferred labels.
Finally, we apply an OOD generalization algorithm on these inferred labels again using 10 different seeds.

\subsection{Hyper-parameter Sampling Grids}
\label{app:hp}

\begin{center}
\begin{tabular}{llll}
\toprule
\textbf{algorithm} & \textbf{hyper-parameter} & \textbf{ResNet}  & \textbf{BERT}\\
\midrule
                   & learning rate & $10^{\text{Uniform}(-5, -3)}$ & $10^{\text{Uniform}(-6, -4)}$\\
\method{}, ERM,          & weight decay  & $10^{\text{Uniform}(-6, -3)}$ & $10^{\text{Uniform}(-6, -3)}$\\
SUBG, RWG          & batch size    & $2^{\text{Uniform}(5, 7)}$    & $2^{\text{Uniform}(4, 6)}$\\
                   & dropout       & ---                           & $\text{Random}([0, 0.1, 0.5])$\\
\midrule
GroupDRO           & $\eta$        & $10^{\text{Uniform}(-3, -1)}$ & $10^{\text{Uniform}(-3, -1)}$\\
\bottomrule
\end{tabular}
\end{center}

\subsection{Statistics of the Datasets}
\vspace{-0.1cm}
\label{app:datasets_stats}
\begin{table}[ht]
  \centering
  \small
  \begin{tabular}{lcccc}
    \toprule
    \textbf{Dataset} & \textbf{Data type} & \textbf{Number of envs.} & \textbf{Number of classes} & \textbf{Dataset size} \\
    \midrule
    Waterbirds & Image & 2 & 2 & 11788 \\
    CelebA & Image & 2 & 2 & 202599 \\
    CivilComments & Text & 8 & 2 & 242436 \\
    MultiNLI & Text & 2 & 3 & 412349 \\
    MetaShift & Image & 2 & 2 & 3499 \\
    ImageNetBG & Image & 2 & 9 & 192255 \\
    VLCS & Image & 4 & 5 & 10729 \\
    PACS & Image & 4 & 7 & 9991 \\
    OfficeHome & Image & 4 & 65 & 15588 \\
    Terra Incognita & Image & 4 & 10 & 24788 \\
    DomainNet & Image & 6 & 345 & 586575 \\
    \bottomrule
  \end{tabular}
\end{table}

\vspace{-0.2cm}
\subsection{Learning to Split on Waterbirds}
\label{app:ls}
We benchmarked the official learning to split code-base \url{https://github.com/YujiaBao/ls} on the WaterBirds dataset. 
We found on a Volta-32GB GPU running the learning to split group inference module took approximately 20 hours. 
We assessed the method's sensitivity to two hyperparameters: the number of epochs used for early stopping (\texttt{patience} argument in the codebase) and the pre-supposed ratio of groups (based on the \texttt{ratio} argument in the code).
For \texttt{patience} we swept over (2, 5, 10) with 5 being the default value. For \texttt{ratio}, we swept over (0.25, 0.5, 0.75) with 0.75 being the default value based on the paper. We found worst group performance using a fixed GroupDRO phase-2 training 
varied by as much as $\pm 7\percent{}$ on Waterbirds.

\vspace{-0.2cm}
\section{\method{} in PyTorch}
\label{app:code}

\definecolor{black}{RGB}{0, 0, 0}
\definecolor{white}{RGB}{255, 255, 255}
\definecolor{light}{RGB}{125, 125, 125}
\definecolor{dark}{RGB}{50, 50, 50}
\definecolor{main}{RGB}{123, 17, 19}
\definecolor{bluem}{RGB}{6, 104, 225}
\definecolor{codegreen}{rgb}{0,0.6,0}
\definecolor{codegray}{rgb}{0.5,0.5,0.5}
\definecolor{codepurple}{rgb}{0.58,0,0.82}
\definecolor{backcolour}{rgb}{0.95,0.95,0.92}

\lstdefinestyle{mystyle}{
    commentstyle=\color{light},
    keywordstyle=\color{main},
    stringstyle=\color{main},
    basicstyle=\ttfamily\scriptsize,
    breakatwhitespace=false,
    breaklines=true,
    captionpos=b,
    keepspaces=true,
    numbers=left,
    numbersep=5pt,
    framexleftmargin=10pt,
    xleftmargin=10pt,
    numberstyle=\tiny\ttfamily\color{light},
    showspaces=false,
    rulecolor=\color{light},
    showstringspaces=false,
    showtabs=false,
    tabsize=2,
    frame=lines,
    escapechar={|}
}

\lstset{style=mystyle}

\begin{lstlisting}[language=Python]
import torch
from torch.nn.functional import cross_entropy

def balanced_cross_entropy(p, y):
    losses = torch.nn.functional.cross_entropy(p, y, reduction="none")
    return sum([losses[y == yi].mean() for yi in y.unique()])

def xrm(x_tr, y_tr, x_va, y_va, lr=1e-2, max_iters=1000):
    # init twins and assign examples (Section 4.1)
    nc = len(y_tr.unique())
    net_a = torch.nn.Linear(x_tr.size(1), nc)
    net_b = torch.nn.Linear(x_tr.size(1), nc)
    net_a.weight.data.mul_(0.0)
    net_b.weight.data.mul_(0.0)
    ind_a = torch.zeros(len(x_tr), 1).bernoulli_(0.5).long()

    # training (Section 4.2)
    opt = torch.optim.SGD(
        list(net_a.parameters()) + list(net_b.parameters()), lr)

    for iteration in range(max_iters):
        pred_a, pred_b = net_a(x_tr), net_b(x_tr)
        pred_hi = pred_a * ind_a + pred_b * (1 - ind_a)
        pred_ho = pred_a * (1 - ind_a) + pred_b * ind_a
        
        opt.zero_grad()
        balanced_cross_entropy(pred_hi, y_tr).backward()
        opt.step()

        # label flipping, useful for model selection (Section 4.3)
        p_ho, y_ho = pred_ho.softmax(dim=1).detach().max(1)
        is_flip = torch.bernoulli((p_ho - 1 / nc) * nc / (nc - 1)).long()
        y_tr = is_flip * y_ho + (1 - is_flip) * y_tr
       
    # environment discovery (Section 4.4) 
    cm = lambda x, y: torch.logical_or(
        net_a(x).argmax(1).ne(y),
        net_b(x).argmax(1).ne(y)).long().detach()
        
    return cm(x_tr, y_tr), cm(x_va, y_va)
\end{lstlisting}

The code above may be helpful to clarify our exposition in the main text.
For an end-to-end example running linear XRM and GroupDRO, see: \url{https://github.com/facebookresearch/XRM/quick_run.py}. 

\end{document}